\documentclass{article}

\usepackage[numbers,sort&compress]{natbib}
\usepackage{PRIMEarxiv}

\usepackage[utf8]{inputenc} 
\usepackage[T1]{fontenc}    
\usepackage{url}            
\usepackage{booktabs}       
\usepackage{amsfonts}       
\usepackage{nicefrac}       
\usepackage{microtype}      
\usepackage{lipsum}
\usepackage{fancyhdr}       
\usepackage{graphicx}       
\graphicspath{{media/}}     

\usepackage[T1]{fontenc}
\usepackage[utf8]{inputenc}
\usepackage{microtype}

\usepackage{graphicx}
\usepackage{svg}
\usepackage{url}

\usepackage{amsmath}
\usepackage{amssymb}
\usepackage{amsfonts}
\usepackage{amsthm}
\usepackage{nicefrac}

\usepackage{booktabs}
\usepackage{tabularx}
\usepackage{array}
\usepackage{ragged2e}
\usepackage{multirow}
\usepackage{makecell}
\usepackage{threeparttable}
\usepackage[table]{xcolor}
\usepackage{rotating}

\usepackage{algorithm}
\usepackage{algpseudocode}

\usepackage{enumitem}
\usepackage{comment}
\usepackage[normalem]{ulem}
\usepackage{xspace}

\usepackage[hidelinks]{hyperref}
\usepackage{cleveref}

\newcolumntype{Y}{>{\RaggedRight\arraybackslash}X}

\newcommand{\smallcaps}[1]{\textsc{#1}\xspace}
\newcommand{\model}{\smallcaps{HARVE}}
\newcommand{\benchmark}{\smallcaps{RewardHackBench}}

\definecolor{ImproveGreen}{HTML}{1B7F1B}
\definecolor{DegradeRed}{HTML}{B22222}
\definecolor{badcell}{RGB}{242,150,150}
\definecolor{midcell}{RGB}{248,190,120}
\definecolor{mildcell}{RGB}{255,238,170}
\definecolor{carow}{RGB}{225,240,252}

\usepackage[most]{tcolorbox}

\title{\model: Hacking-Aware Reward-Head Vector Editing for Robust Reward Models}

\author{
  Shuang Liu \\
  Carnegie Mellon University \\
  \And
  Yuxuan Bo \\
  University of Virginia \\
  \And
  Qiuyang Zhao \\
  Harvard University \\
  \And
  Caiyue Huang \\
  Stanford University \\
  \And
  Xiaorong Chen \\
  University of Michigan \\
  \And 
  Yanguang Liu \\
  New Jersey Institute of Technology\\
  \And
  Mengnan Du \\
  The Chinese University of Hong Kong, Shenzhen \\
}

\begin{document}
\maketitle

\begin{abstract}
Reward models are central to large language model (LLM) alignment, but they remain vulnerable to reward hacking.
To evaluate reward-model robustness, we introduce \benchmark with 1,203 matched gold--hacked pairs. \benchmark spans 13 reward-hacking patterns covering real-world high-stakes domains and general settings, and we find severe failures on specific subcategories across eight reward models. 
To mitigate these failures, we propose \model, a training-free reward-head editing method for scalar reward models. Instead of fine-tuning the reward model, \model identifies a multi-directional hacking subspace from residual-stream directions associated with selected hacking subcategories, and removes the component of the reward-head vector aligned with that subspace. This directly reduces the reward head's sensitivity to hacking-related features using only a small set of contrastive gold--hacked examples, without gradient updates or fine-tuning. Comprehensive experiments across eight reward models indicate that \model improves hacking robustness, outperforms fine-tuning baselines, and preserves general reward-model capability. Further analyses suggest that reward hacking is better captured as a multidimensional residual-space structure than by isolated surface cues. The code is available at \url{https://github.com/olivialiu121/HARVE-Reward-Head-Editing}.
\end{abstract}

\section{Introduction}
\label{sec:introduction}

Modern LLM alignment often relies on reward models to score and improve model outputs \cite{Stiennon2020, ouyang2022training}. However, reward models are vulnerable to \emph{reward hacking}: models may exploit reward-correlated cues such as length or stylistic polish without improving factual correctness or reasoning quality \citep{singhal2023long, gao2023scaling, park2024offsetbias, liu2025rm}.
Existing benchmarks study generic reward-model hacking such as length bias and formatting bias \cite{singhal2023long, park2024offsetbias, sharma2024sycophancy}, with recent extensions to chat, code, math, and safety \citep{lambert-etal-2025-rewardbench, liu2025rm}. However, reward hacking in complex high-stakes domains such as law and compliance in real-world settings remains underexplored.

To fill this gap, we introduce \benchmark, a benchmark with 1,203 matched gold--hacked pairs spanning 13 subcategories. \benchmark covers both professional-domain attacks and general-purpose attacks. Evaluating eight open-source scalar reward models on \benchmark, we find severe failures in specific subcategories, motivating mitigation methods that directly address reward hacking.

Current mitigation methods typically collect debiased or adversarial preference data and fine-tune the reward model \citep{shen2023trickle, liu2025rrm, chen2024odin}. These approaches are data-dependent, computationally costly, and difficult to repeat for each newly discovered hacking pattern. 
To this end, we propose \model{}, a training-free reward-head editing method for scalar reward models. 
Instead of fine-tuning the reward model, \model{} identifies a multi-directional hacking subspace from residual-stream directions associated with selected hacking subcategories, and removes the component of the reward-head vector $w_r$ aligned with that subspace. This reduces the reward head's sensitivity to hacking-related features using only a small set of contrastive gold--hacked examples, without gradient updates or fine-tuning.

\begin{table*}[!ht]
\centering
\scriptsize
\setlength{\tabcolsep}{2.5pt}
\renewcommand{\arraystretch}{1.2} 
\caption{
\textbf{Baseline reward-model performance on the held-out test split of \benchmark{}.}
Each cell reports the gold-preference rate, i.e., the percentage of pairs where the reward model assigns a higher score to the gold response than to the hacked response; higher is better.
A1--C3 are professional-domain subcategories, and D1--E3 are general-purpose subcategories from LLMBar~\citep{zeng2024llmbar} via RewardBench~\citep{lambert-etal-2025-rewardbench}.
Full subcategory names are provided in Table~\ref{tab:taxonomy}.
\textbf{Cell shading:}
\colorbox{badcell}{red} $\leq30\%$ severe failure;
\colorbox{midcell}{orange} $30$--$50\%$ below chance;
\colorbox{mildcell}{yellow} $50$--$70\%$ mediocre;
white $\geq70\%$.
\emph{Avg.} is the sample-weighted micro-average over all 13 subcategories of the test split ($N=967$).
}
\label{tab:results_scalar}
\resizebox{\textwidth}{!}{%
\begin{tabular}{llcccccccccccccc}
\toprule
\multirow{3}{*}{\textbf{Reward Model}} & \multirow{3}{*}{\textbf{Size}}
  & \multicolumn{8}{c}{\textit{Professional-domain categories}}
  & \multicolumn{5}{c}{\textit{General-purpose categories}}
  & \multirow{3}{*}{\textbf{Avg.}} \\
\cmidrule(lr){3-10}\cmidrule(lr){11-15}
  & & \multicolumn{3}{c}{\textbf{A. Surface}}
    & \multicolumn{2}{c}{\textbf{B. Reasoning}}
    & \multicolumn{3}{c}{\textbf{C. Sycophantic}}
    & \multicolumn{2}{c}{\textbf{D. Off-Topic}}
    & \multicolumn{3}{c}{\textbf{E. Style}}
    & \\
\cmidrule(lr){3-5}\cmidrule(lr){6-7}\cmidrule(lr){8-10}\cmidrule(lr){11-12}\cmidrule(lr){13-15}
  & & \textbf{A1} & \textbf{A2} & \textbf{A3}
    & \textbf{B1} & \textbf{B2}
    & \textbf{C1} & \textbf{C2} & \textbf{C3}
    & \textbf{D1} & \textbf{D2}
    & \textbf{E1} & \textbf{E2} & \textbf{E3} & \\
\midrule
Skywork-Reward-V2-Qwen3      & 0.6B & \cellcolor{midcell}33.87  & 86.96                     & \cellcolor{midcell}46.27  & \cellcolor{mildcell}58.57 & 88.57                     & 94.29                     & 82.86                     & 71.43                     & \cellcolor{mildcell}58.21 & \cellcolor{midcell}43.48  & \cellcolor{mildcell}63.83 & 89.00                     & \cellcolor{mildcell}58.70 & \cellcolor{mildcell}\textbf{67.53} \\
InternLM2-reward             & 1.8B & 93.55                     & 81.16                     & 95.52                     & \cellcolor{midcell}45.71  & 87.14                     & \cellcolor{mildcell}51.43 & \cellcolor{mildcell}64.29 & 74.29                     & \cellcolor{mildcell}58.21 & \cellcolor{midcell}45.65  & 80.85                     & 88.00                     & \cellcolor{mildcell}58.70 & \textbf{70.01} \\
GRM-Llama-3.2                & 3B   & \cellcolor{midcell}43.55  & \cellcolor{mildcell}66.67 & \cellcolor{mildcell}53.73 & \cellcolor{mildcell}67.14 & 100.00                    & 74.29                     & 98.57                     & \cellcolor{badcell}21.43  & 85.82                     & 93.48                     & 74.47                     & 88.00                     & \cellcolor{mildcell}69.57 & \textbf{74.15} \\
Skywork-Reward-V2-Llama-3.2  & 3B   & 85.48                     & 84.06                     & \cellcolor{mildcell}65.67 & 78.57                     & 100.00                    & 90.00                     & 80.00                     & 75.71                     & 79.10                     & 85.87                     & \cellcolor{mildcell}65.96 & 95.00                     & 82.61                     & \textbf{82.83} \\
RM-Mistral                   & 7B   & 90.32                     & 85.51                     & 76.12                     & \cellcolor{mildcell}58.57 & 100.00                    & 80.00                     & 98.57                     & \cellcolor{mildcell}57.14 & \cellcolor{midcell}44.78  & \cellcolor{mildcell}51.09 & \cellcolor{mildcell}61.70 & 89.00                     & \cellcolor{midcell}43.48  & \textbf{71.04} \\
Skywork-Reward-Llama-3.1     & 8B   & 85.48                     & \cellcolor{mildcell}60.87 & 71.64                     & \cellcolor{mildcell}68.57 & 100.00                    & 98.57                     & 95.71                     & \cellcolor{mildcell}62.86 & 90.30                     & 95.65                     & 80.85                     & 91.00                     & 73.91                     & \textbf{84.07} \\
FsfairX-LLaMA3-RM            & 8B   & 83.87                     & 84.06                     & 76.12                     & \cellcolor{mildcell}57.14 & 100.00                    & 90.00                     & 92.86                     & \cellcolor{midcell}40.00  & \cellcolor{mildcell}50.75 & \cellcolor{mildcell}57.61 & \cellcolor{mildcell}65.96 & 91.00                     & \cellcolor{mildcell}52.17 & \textbf{71.77} \\
InternLM2-reward             & 20B  & 98.39                     & 75.36                     & 79.10                     & \cellcolor{mildcell}61.43 & 100.00                    & 91.43                     & 80.00                     & 77.14                     & \cellcolor{mildcell}64.93 & \cellcolor{mildcell}68.48 & 80.85                     & 92.00                     & 86.96                     & \textbf{79.94} \\
\bottomrule
\end{tabular}
}
\end{table*}

We evaluate \model on both \benchmark and RM-Bench \citep{liu2025rm}, comparing it with fine-tuning baselines. Across eight reward models, \model achieves the best \benchmark performance, improving gold-preference rate on target subcategories by 21.1 percentage points over the original reward model baseline and 13.7 points over the stronger fine-tuning baseline. It also largely preserves performance on non-target subcategories of \benchmark and improves RM-Bench performance by 2.4 points. We further conduct ablation experiments to show that the edited directions capture presentation patterns beyond literal length and provide evidence that reward hacking is represented as a multidimensional subspace in reward-model residual representations.

Our contributions in this work are threefold. First, we propose \model, an efficient, interpretable, and training-free mitigation approach via reward-head editing. Second, we introduce \benchmark, covering reward hacking in both general-purpose and real-world high-stakes domains such as law and compliance. Third, we provide empirical and mechanistic evidence that reward hacking is represented as a multidimensional subspace in reward-model residual representations, enabling targeted mitigation through vector editing.

\section{\benchmark}
\label{sec:benchmark}

To evaluate reward-model robustness against reward hacking in both professional-domain and general-purpose settings, we introduce \benchmark, a benchmark of 1,203 matched gold--hacked response pairs across 13 subcategories. The benchmark covers professional-domain attacks in law, policy, and compliance as well as general-purpose attacks. For each query, the hacked response preserves the gold response's length, structure, and register while introducing one targeted hacking pattern. We split the professional-domain categories into stratified 20\%/10\%/70\% train/dev/test partitions and reserve all general-purpose examples for testing, yielding a held-out test split of 967 pairs: 548 professional-domain pairs and 419 general-purpose pairs.

\subsection{Benchmark Construction}
\benchmark contains 13 subcategories grouped into five reward-hacking categories: 
(A) \emph{Surface-Form Mimicry (Surface)}, where responses use fake or unsupported authority; 
(B) \emph{Broken Reasoning (Reasoning)}, where responses omit required reasoning elements or invert key burdens; 
(C) \emph{Sycophantic Hacking (Sycophantic)}, where responses over-align with the user or remove appropriate caveats; 
(D) \emph{Off-Topic Hacking (Off-Topic)}, where responses answer a related but different prompt; and 
(E) \emph{Style-Over-Substance (Style)}, where polished writing masks factual, logical, or task-specific flaws. 
Categories A--C target professional-domain settings such as law, policy, and compliance, while D--E cover general-purpose attacks. The full 13-subcategory taxonomy is provided in Appendix~\ref{app:taxonomy}.

\begin{figure*}[!ht]
    \centering
    \includegraphics[width=\textwidth]{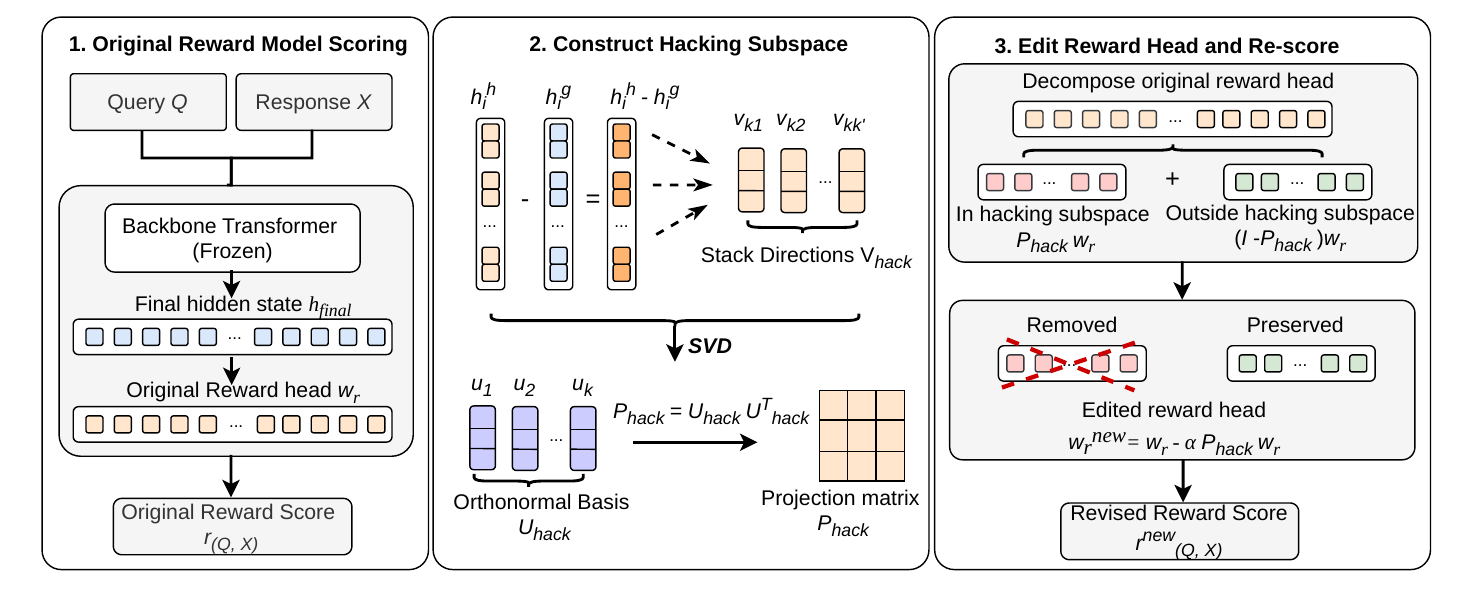}
    \caption{Overview of \model{}. 
    Given a scalar reward model, we first identify training pairs where the RM prefers the hacked response over the gold response. 
    We use these fooled pairs to estimate residual-stream directions associated with reward-hacking patterns. 
    \model{} then edits the reward-head vector by removing its projection onto the subspace spanned by these hacking directions, leaving the transformer backbone unchanged.
    }
    \label{fig:method-overview}
\end{figure*}

For the professional-domain portion, we construct 784 matched gold--hacked pairs using a human--LLM pipeline. Two New York Bar-admitted attorneys draft or revise questions and gold responses using LLM-written examples and samples from existing legal benchmarks, such as \textsc{LegalBench}, \textsc{LEXam}, and \textsc{Stanford Bar Exam QA}~\citep{guha2023legalbench, lexam2024, zheng2025barexam}. Then we ask GPT-5-mini to rewrite each gold response into a hacked response that preserves the length and structure of the gold response while introducing one targeted failure mode. All pairs are then reviewed by three U.S. law students under attorney supervision. 
The final professional-domain subset has a mean hacked-to-gold word-count ratio of $1.01$, reducing length and formatting confounds~\citep{singhal2023long, park2024offsetbias}. 
For the general-purpose portion, we include 419 gold--hacked pairs from LLMBar via the Chat Hard subset of \textsc{RewardBench}~\citep{zeng2024llmbar, lambert-etal-2025-rewardbench}.

\subsection{Evaluation Protocol}
We evaluate eight open-source scalar reward models spanning 0.6B--20B parameters from the Skywork, InternLM2, GRM, RM-Mistral, and FsfairX families~\citep{liu2025skywork, cai2024internlm2, yang2024GRM, dong2023raft, xiong2024iterative, dong2024rlhf}. See Appendix \ref{app:rm-details} for detailed information. 
For each prompt $x$ with gold response $y_g$ and hacked response $y_h$, we report the \emph{gold-preference rate}: the fraction of pairs where $R(x,y_g)>R(x,y_h)$. 
Higher values indicate stronger robustness to reward hacking; ties are counted as non-gold preferences.

\subsection{Benchmark Results}
\Cref{tab:results_scalar} shows that open-source reward models remain vulnerable to reward hacking. 
Across eight scalar reward models, average gold-preference rates range from 67.53\% to 84.07\%, but every model has at least one mediocre-or-worse subcategory and several fall below chance on specific attacks. For instance, \textit{C3 (Hedge Stripping)} yields severe failures, including 21.43\% for GRM-Llama-3.2-3B and 40.00\% for FsfairX-LLaMA3-RM-8B. 
Model scale alone does not solve the problem. For example, InternLM2-reward-20B underperforms several smaller reward models such as Skywork-Reward-V2-Llama-3.2-3B and Skywork-Reward-Llama-3.1-8B on most subcategories. 
Overall, \benchmark{} shows that reward hacking is heterogeneous and category-specific, motivating mitigation methods that directly target diverse hacking directions.

\section{Mitigation Method} 
\label{sec:carve-method}

We propose \model, a training-free method for mitigating reward hacking in scalar reward models. 
\model estimates residual-stream directions associated with hacking patterns from gold--hacked examples, then edits the reward-head vector by removing its projection onto the corresponding subspace. This reduces sensitivity to hacking-related features without gradient updates, fine-tuning, or additional inference-time cost.

\subsection{Reward-Head Linearity}
\label{sec:carve-linearity}

We focus on scalar reward models, which assign a single real-valued score to each prompt--response pair. 
Given final hidden state $h_{\mathrm{final}}(x,y)\in\mathbb{R}^d$, the reward is computed by a linear reward head:
$r(x,y)=w_r^\top h_{\mathrm{final}}(x,y)$,
where $w_r$ is the reward-head vector. 
Because the score depends linearly on $w_r$, reducing sensitivity to a hacking-related direction can be formulated as a geometric edit to the reward head.

\begin{algorithm}[t]
\caption{\model{}: Reward-Head Editing}
\label{alg:harve}
\begin{algorithmic}[1]
\Require Reward head $w_r$; training pairs $\mathcal{D}_k = \{(x_i, y_i^g, y_i^h)\}$ for target subcategories $k \in \mathcal{K}$; edit strength $\alpha$
\Ensure Edited reward head $w_r^{\mathrm{new}}$

\For{$k \in \mathcal{K}$}
    \State Identify fooled subset: \\
    $F_k \gets \{i \in \{1,\ldots,N_k\} \mid r(x_i, y_i^h) > r(x_i, y_i^g)\}$
    \State Extract hacking direction: \\
    $v_k \leftarrow \text{Normalize}\big(\sum_{i \in F_k} (h_i^h - h_i^g)\big)$
\EndFor

\State Stack subcategory directions: $V_{\mathrm{hack}} \leftarrow [v_{k_1}, \dots, v_{k_K}]$
\State Compute projection matrix via SVD: \\
    $P_{\mathrm{hack}} \leftarrow U U^\top \text{ where } U = \text{SVD}(V_{\mathrm{hack}})$
\State Remove hacking components and \Return \\
    $w_r^{\mathrm{new}} \leftarrow w_r - \alpha P_{\mathrm{hack}} w_r$
\end{algorithmic}
\end{algorithm}

\subsection{Estimating Hacking Directions}
\label{sec:carve-extraction}

For each reward model, we select a small set of \emph{target subcategories} whose hacking directions are estimated and removed. 
All remaining subcategories are treated as \emph{non-target} categories and used to evaluate whether the edit preserves unrelated reward-model behavior.

For each target subcategory $k$, let
$\mathcal{D}_k=\{(x_i,y_i^g,y_i^h)\}_{i=1}^{N_k}$
denote the matched training pairs, where $y_i^g$ is the gold response and $y_i^h$ is the hacked response. 
We first identify the fooled subset, consisting of examples where the base reward model prefers the hacked response:
$F_k
=
\{\, i\in\{1,\ldots,N_k\} :
r(x_i,y_i^h)>r(x_i,y_i^g) \,\}$.
For each $i\in F_k$, we extract the final hidden states
$h_i^g = h_{\mathrm{final}}(x_i,y_i^g)$ and
$h_i^h = h_{\mathrm{final}}(x_i,y_i^h)$.
The hacking direction for subcategory $k$ is the normalized mean hidden-state shift from gold to hacked responses over the fooled subset:
\begin{equation}
\begin{aligned}
\tilde v_k
&=
\frac{1}{|F_k|}
\sum_{i\in F_k}
\left(h_i^h-h_i^g\right), \\
v_k
&=
\frac{\tilde v_k}{\lVert \tilde v_k\rVert}.
\end{aligned}
\label{eq:hacking-direction}
\end{equation}
Thus, $v_k$ captures the residual-stream direction associated with successful reward hacking in subcategory $k$. 
If $F_k$ is empty, we omit that subcategory for the corresponding reward model.

\subsection{Constructing the Hacking Subspace}
\label{sec:carve-subspace}

Let $\mathcal{K}=\{k_1,\ldots,k_K\}$ denote the target subcategories retained for a reward model. 
We stack their directions into the hacking-direction matrix
$V_{\mathrm{hack}}
=
[v_{k_1},\ldots,v_{k_K}]
\in \mathbb{R}^{d\times K}$.
The corresponding hacking subspace is the column space of this matrix,
$\mathcal{S}_{\mathrm{hack}}
= \operatorname{span}(V_{\mathrm{hack}})
= \operatorname{span}\{v_{k_1},\ldots,v_{k_K}\}$.

Because the directions $v_k$ are not generally orthogonal and may overlap across hacking patterns, we compute an orthonormal basis for their span using singular value decomposition (SVD). 
Specifically, we decompose
\begin{equation}
V_{\mathrm{hack}} = U\Sigma Q^\top,
\label{eq:hacking-svd}
\end{equation}
where the columns of $U$ are orthonormal directions in the residual-stream space. 
Let $\sigma_j$ denote the singular values on the diagonal of $\Sigma$. 
We retain the columns of $U$ with $\sigma_j>10^{-6}$ and denote the resulting basis by $U_{\mathrm{hack}}$. 
This threshold is used only as a numerical safeguard against degenerate or exactly collinear directions, not as an aggressive rank-reduction procedure. 
The projection matrix onto the hacking subspace is
\begin{equation}
P_{\mathrm{hack}}
=
U_{\mathrm{hack}}U_{\mathrm{hack}}^\top .
\label{eq:hacking-projection}
\end{equation}

\subsection{Reward-Head Subspace Editing}
\label{sec:carve-editing}

Given the projection matrix $P_{\mathrm{hack}}$, \model{} edits the reward head by subtracting the component of $w_r$ aligned with the hacking subspace:
\begin{equation}
w_r^{\mathrm{new}}
=
w_r-\alpha P_{\mathrm{hack}}w_r,
\label{eq:carve-update}
\end{equation}
where $\alpha\ge0$ controls the intervention strength. 
When $\alpha=0$, the reward head is unchanged; when $\alpha=1$, the hacking-aligned component of $w_r$ is removed. 
Values $\alpha>1$ subtract more than this component, applying a stronger correction. 
We select $\alpha$ on the development split using validation trade-off curves.
After editing, the reward model scores examples as
$r_{\mathrm{new}}(x,y)
=
(w_r^{\mathrm{new}})^\top h_{\mathrm{final}}(x,y)$.

\begin{table*}[!htb]
\centering
\scriptsize
\setlength{\tabcolsep}{2.5pt}
\caption{
\textbf{Main evaluation of \model{} against fine-tuning baselines.}
For each reward model, we compare the original reward model, 3:1 and 5:1 data-augmented fine-tuning baselines, and \model at the selected edit strength $\alpha^\star$.
On \benchmark, \textbf{Tgt} and \textbf{Non-Tgt} report average gold-preference rates over the subcategories used for editing and the remaining subcategories, respectively; Surface, Reasoning, Sycophantic, Off-Topic, and Style denote the five top-level \benchmark categories.
On RM-Bench Hard, \textbf{Avg} averages gold-preference rate over chat (Cht), code (Cde), math (Mth), safety-refuse (S-Rf), and safety-response (S-Rs). 
Bold values mark the best non-baseline method within each block, and blue rows highlight \model. 
$\uparrow$ Higher is better.
}
\label{tab:methods-aggregate-absolute-sweet}
\resizebox{\textwidth}{!}{%
\begin{tabular}{lrrrrrrrrrrrrr}
\toprule
& \multicolumn{7}{c}{\textbf{\benchmark $\uparrow$}} & \multicolumn{6}{c}{\textbf{RM-Bench $\uparrow$}} \\
\cmidrule(lr){2-8}\cmidrule(lr){9-14}
& \multicolumn{2}{c}{\textit{Overall}} & \multicolumn{5}{c}{\textit{By category}} & \multicolumn{1}{c}{\textit{Overall}} & \multicolumn{5}{c}{\textit{By domain}} \\
\cmidrule(lr){2-3}\cmidrule(lr){4-8}\cmidrule(lr){9-9}\cmidrule(lr){10-14}
\textbf{Model} & \textbf{Tgt} & \textbf{Non-Tgt} & \textbf{Surface} & \textbf{Reasoning} & \textbf{Sycophantic} & \textbf{Off-Topic} & \textbf{Style} & \textbf{Avg} & \textbf{Cht} & \textbf{Cde} & \textbf{Mth} & \textbf{S-Rf} & \textbf{S-Rs} \\
\midrule
\textbf{Skywork-V2-Qwen3 (0.6B)} & & & & & & & & & & & & & \\
\quad + Baseline  & $53.16$ & $73.07$ & $56.57$ & $73.57$ & $82.86$ & $52.21$ & $75.65$ & $56.90$ & $26.36$ & $44.15$ & $54.32$ & $88.97$ & $51.17$ \\
\quad + Finetune (3:1) & $58.36$ & $74.93$ & $60.61$ & $75.71$ & $85.71$ & $56.64$ & $75.65$ & $57.17$ & $26.87$ & $43.57$ & $54.00$ & $88.85$ & $55.20$ \\
\quad + Finetune (5:1) & $58.36$ & $75.50$ & $61.62$ & $75.00$ & $86.19$ & $57.52$ & $75.65$ & $57.00$ & $25.32$ & $\mathbf{43.86}$ & $53.94$ & $88.50$ & $55.41$ \\
\rowcolor{carow}
\quad + \textbf{\model ($\alpha^\star{=}1.25$, ours)} & $\mathbf{85.50}$ & $\mathbf{77.51}$ & $\mathbf{80.81}$ & $\mathbf{82.14}$ & $\mathbf{92.38}$ & $\mathbf{69.03}$ & $\mathbf{75.65}$ & $\mathbf{62.04}$ & $\mathbf{49.87}$ & $37.13$ & $\mathbf{57.66}$ & $\mathbf{96.01}$ & $\mathbf{61.57}$ \\
\midrule
\textbf{InternLM2-reward (1.8B)} & & & & & & & & & & & & & \\
\quad + Baseline  & $58.93$ & $74.24$ & $89.90$ & $66.43$ & $63.33$ & $52.21$ & $79.27$ & $59.56$ & $24.29$ & $25.15$ & $69.69$ & $92.61$ & $44.59$ \\
\quad + Finetune (3:1) & $61.07$ & $74.82$ & $90.40$ & $70.00$ & $65.24$ & $52.65$ & $78.76$ & $59.33$ & $\mathbf{24.03}$ & $24.56$ & $\mathbf{69.44}$ & $92.61$ & $44.59$ \\
\quad + Finetune (5:1) & $62.50$ & $75.25$ & $\mathbf{90.91}$ & $68.57$ & $67.14$ & $53.98$ & $79.27$ & $58.96$ & $23.51$ & $23.83$ & $68.75$ & $\mathbf{92.84}$ & $\mathbf{44.80}$ \\
\rowcolor{carow}
\quad + \textbf{\model ($\alpha^\star{=}1.10$, ours)} & $\mathbf{78.93}$ & $\mathbf{75.40}$ & $87.88$ & $\mathbf{78.57}$ & $\mathbf{83.81}$ & $\mathbf{55.31}$ & $\mathbf{79.79}$ & $\mathbf{59.48}$ & $19.90$ & $\mathbf{37.43}$ & $68.24$ & $90.14$ & $39.07$ \\
\midrule
\textbf{GRM-Llama-3.2 (3B)} & & & & & & & & & & & & & \\
\quad + Baseline  & $46.47$ & $84.81$ & $55.05$ & $83.57$ & $64.76$ & $88.94$ & $79.79$ & $53.03$ & $31.78$ & $26.17$ & $36.23$ & $97.30$ & $85.99$ \\
\quad + Finetune (3:1) & $50.56$ & $85.96$ & $59.09$ & $85.00$ & $68.57$ & $88.94$ & $\mathbf{80.31}$ & $53.15$ & $32.82$ & $26.46$ & $36.17$ & $97.42$ & $\mathbf{85.77}$ \\
\quad + Finetune (5:1) & $54.28$ & $\mathbf{86.25}$ & $61.62$ & $83.57$ & $72.86$ & $88.94$ & $80.31$ & $53.18$ & $32.82$ & $26.46$ & $\mathbf{36.23}$ & $97.42$ & $85.77$ \\
\rowcolor{carow}
\quad + \textbf{\model ($\alpha^\star{=}1.25$, ours)} & $\mathbf{82.90}$ & $84.38$ & $\mathbf{72.73}$ & $\mathbf{90.71}$ & $\mathbf{89.52}$ & $\mathbf{90.27}$ & $77.20$ & $\mathbf{56.59}$ & $\mathbf{64.34}$ & $\mathbf{47.08}$ & $28.86$ & $\mathbf{97.89}$ & $82.80$ \\
\midrule
\textbf{Skywork-V2-Llama-3.2 (3B)} & & & & & & & & & & & & & \\
\quad + Baseline & $75.09$ & $85.51$ & $78.28$ & $89.29$ & $81.90$ & $81.42$ & $83.94$ & $69.43$ & $55.04$ & $51.90$ & $62.51$ & $97.54$ & $79.19$ \\
\quad + Finetune (3:1) & $76.17$ & $86.23$ & $79.29$ & $89.29$ & $84.29$ & $81.42$ & $\mathbf{84.46}$ & $69.40$ & $55.30$ & $\mathbf{52.19}$ & $62.26$ & $97.54$ & $79.19$ \\
\quad + Finetune (5:1) & $76.90$ & $86.09$ & $80.30$ & $89.29$ & $84.29$ & $81.42$ & $83.94$ & $69.33$ & $55.04$ & $51.90$ & $62.26$ & $97.54$ & $79.19$ \\
\rowcolor{carow}
\quad + \textbf{\model ($\alpha^\star{=}1.25$, ours)} & $\mathbf{92.42}$ & $\mathbf{86.52}$ & $\mathbf{86.36}$ & $\mathbf{93.57}$ & $\mathbf{96.67}$ & $\mathbf{82.74}$ & $83.42$ & $\mathbf{70.91}$ & $\mathbf{64.34}$ & $50.58$ & $\mathbf{62.76}$ & $\mathbf{98.94}$ & $\mathbf{82.59}$ \\
\midrule
\textbf{RM-Mistral (7B)} & & & & & & & & & & & & & \\
\quad + Baseline  & $67.87$ & $72.32$ & $83.84$ & $79.29$ & $78.57$ & $47.35$ & $71.50$ & $44.41$ & $16.54$ & $25.29$ & $30.88$ & $88.62$ & $60.72$ \\
\quad + Finetune (3:1) & $84.12$ & $71.30$ & $94.44$ & $81.43$ & $\mathbf{93.81}$ & $45.13$ & $64.77$ & $45.74$ & $19.12$ & $26.46$ & $30.75$ & $93.31$ & $60.08$ \\
\quad + Finetune (5:1) & $84.48$ & $\mathbf{78.55}$ & $\mathbf{94.95}$ & $82.14$ & $93.33$ & $57.52$ & $\mathbf{76.17}$ & $47.88$ & $24.55$ & $26.61$ & $33.14$ & $\mathbf{95.19}$ & $62.00$ \\
\rowcolor{carow}
\quad + \textbf{\model ($\alpha^\star{=}2.50$, ours)} & $\mathbf{90.25}$ & $76.52$ & $88.38$ & $\mathbf{92.14}$ & $87.62$ & $\mathbf{65.49}$ & $73.58$ & $\mathbf{65.06}$ & $\mathbf{53.75}$ & $\mathbf{49.27}$ & $\mathbf{52.93}$ & $94.25$ & $\mathbf{85.35}$ \\
\midrule
\textbf{FsfairX-LLaMA3-RM (8B)} & & & & & & & & & & & & & \\
\quad + Baseline  & $57.49$ & $75.66$ & $81.31$ & $78.57$ & $74.29$ & $53.54$ & $75.65$ & $50.46$ & $20.67$ & $37.13$ & $39.19$ & $91.90$ & $57.32$ \\
\quad + Finetune (3:1) & $67.63$ & $77.89$ & $85.86$ & $80.71$ & $83.81$ & $55.75$ & $76.17$ & $50.94$ & $21.71$ & $37.13$ & $39.63$ & $92.96$ & $57.11$ \\
\quad + Finetune (5:1) & $71.50$ & $\mathbf{80.13}$ & $\mathbf{89.90}$ & $82.14$ & $85.71$ & $58.41$ & $\mathbf{78.76}$ & $52.47$ & $24.55$ & $39.33$ & $\mathbf{40.77}$ & $95.19$ & $56.69$ \\
\rowcolor{carow}
\quad + \textbf{\model ($\alpha^\star{=}1.75$, ours)} & $\mathbf{86.96}$ & $76.58$ & $80.81$ & $\mathbf{94.29}$ & $\mathbf{86.19}$ & $\mathbf{62.39}$ & $76.68$ & $\mathbf{54.38}$ & $\mathbf{31.27}$ & $\mathbf{43.86}$ & $40.58$ & $\mathbf{96.13}$ & $\mathbf{59.66}$ \\
\midrule
\textbf{Skywork-Reward-Llama-3.1 (8B)} & & & & & & & & & & & & & \\
\quad + Baseline  & $65.94$ & $91.17$ & $72.22$ & $84.29$ & $85.71$ & $92.04$ & $84.46$ & $54.68$ & $32.56$ & $32.75$ & $36.80$ & $98.36$ & $85.99$ \\
\quad + Finetune (3:1) & $72.46$ & $91.75$ & $76.77$ & $85.71$ & $91.43$ & $\mathbf{91.59}$ & $84.46$ & $54.71$ & $33.33$ & $32.60$ & $36.61$ & $98.36$ & $86.41$ \\
\quad + Finetune (5:1) & $73.55$ & $\mathbf{92.19}$ & $77.78$ & $85.71$ & $92.38$ & $91.59$ & $\mathbf{85.49}$ & $54.79$ & $33.59$ & $32.60$ & $36.67$ & $98.36$ & $86.62$ \\
\rowcolor{carow}
\quad + \textbf{\model ($\alpha^\star{=}1.10$, ours)} & $\mathbf{84.42}$ & $89.58$ & $\mathbf{80.30}$ & $\mathbf{93.57}$ & $\mathbf{96.19}$ & $91.15$ & $79.79$ & $\mathbf{61.22}$ & $\mathbf{50.90}$ & $\mathbf{45.18}$ & $\mathbf{43.04}$ & $\mathbf{98.59}$ & $\mathbf{86.62}$ \\
\midrule
\textbf{InternLM2-reward (20B)} & & & & & & & & & & & & & \\
\quad + Baseline  & $73.19$ & $82.78$ & $83.84$ & $80.71$ & $82.86$ & $66.81$ & $88.08$ & $62.65$ & $29.20$ & $35.82$ & $67.86$ & $95.66$ & $51.80$ \\
\quad + Finetune (3:1) & $73.55$ & $\mathbf{83.07}$ & $84.34$ & $80.71$ & $83.81$ & $\mathbf{66.81}$ & $\mathbf{88.08}$ & $62.57$ & $29.20$ & $\mathbf{35.96}$ & $\mathbf{67.61}$ & $95.66$ & $51.80$ \\
\quad + Finetune (5:1) & $75.36$ & $83.07$ & $84.85$ & $83.57$ & $83.81$ & $66.81$ & $88.08$ & $62.47$ & $28.68$ & $35.67$ & $67.36$ & $95.77$ & $52.44$ \\
\rowcolor{carow}
\quad + \textbf{\model ($\alpha^\star{=}1.60$, ours)} & $\mathbf{88.41}$ & $80.90$ & $\mathbf{92.42}$ & $\mathbf{90.71}$ & $\mathbf{89.05}$ & $63.72$ & $83.42$ & $\mathbf{63.78}$ & $\mathbf{35.40}$ & $35.53$ & $66.98$ & $\mathbf{95.89}$ & $\mathbf{59.24}$ \\
\bottomrule
\end{tabular}
}
\end{table*}

\section{Experiments}

\subsection{Setups}
\label{sec:experiments-setup}

\paragraph{Baselines.}
We compare \model with data-augmented fine-tuning baselines that fine-tune the reward model on a mixture of general preference data and pairs from the \benchmark training set. 
We consider two mixture ratios, 3:1 and 5:1, where the ratio denotes the number of general preference pairs per \benchmark pair. Fine-tuning details are provided in Appendix~\ref{app:ft-details}.

\paragraph{Data splits.}
For the professional-domain categories, we use stratified 20\%/10\%/70\% train/dev/test splits. 
The general-purpose categories are reserved entirely for held-out testing. The extraction split is used to estimate \model{}'s hacking directions and to provide benchmark-specific data for the fine-tuning baselines. 
The development split is used only to select the intervention strength $\alpha^\star$, and the test split is used for all reported results.  We use the same subcategory-stratified split for every reward model.

\paragraph{\model vector extraction.}
For each reward model, we select three or four lowest-performing professional-domain subcategories as \emph{target subcategories} using only the training split. We then estimate the hacking direction $v_k$ from fooled examples for each selected target subcategory. 

\paragraph{Benchmarks and Metrics.} We evaluate the eight reward models on \benchmark{} and additionally use RM-Bench~\citep{liu2025rm} to measure cross-domain transfer across chat, code, math, and safety. RM-Bench contains three difficulty-level subsets, easy, normal, hard~\citep{liu2025rm}, and we evaluate on the hard set.
For each prompt $x$ with gold response $y_g$ and hacked response $y_h$, a scalar reward model $R$ assigns scores $R(x,y_g)$ and $R(x,y_h)$. We report the \emph{gold-preference rate}, defined as the fraction of pairs for which $R(x,y_g) > R(x,y_h)$.

\subsection{Main Results}
\label{sec:main-results}

For each reward model, we compare \model with the two fine-tuning baselines, and report performance on targeted subcategories of \benchmark, non-target subcategories of \benchmark, and RM-Bench Hard task sets \cite{liu2025rm}. 
The editing strength $\alpha^\star$ is selected per reward model by choosing the value that maximizes the performance on \emph{target subcategories} while balancing the performance on RM-Bench. See \Cref{tab:methods-aggregate-absolute-sweet} for the full experiment results.

\paragraph{\model substantially improves robustness in target subcategories.}
\model achieves the best performance on target subcategories of \benchmark for all eight reward models. On average, it improves the gold-preference rate of target subcategories by +21.1 percentage points over the original reward model baseline and by +13.7 points over the stronger fine-tuning baseline. The gains are consistent across model families and sizes, ranging from +10.5 percentage points for Skywork-Reward-Llama-3.1-8B to +32.3 points for Skywork-V2-Qwen3-0.6B.
\model{} improves performance on non-target subcategories over the original reward model for all eight RMs. Although fine-tuning baselines sometimes achieve higher gold-preference rates on individual non-target subcategories, \model{} produces substantially larger gains on target subcategories while keeping non-target performance stable. 
These results suggest that reward-head vector editing can remove hacking-sensitive directions more effectively than fine-tuning, without broadly degrading unrelated categories.

\paragraph{\model transfers to cross-domain reward-hacking settings.}
\model also preserves or improves RM-Bench performance across chat, code, math, and safety domains. Across the eight reward models, \model generally preserves or improves RM-Bench performance. For example, on RM-Mistral-7B, \model improves RM-Bench average performance by +20.65 percentage points over the original reward model and by +17.18 points over the stronger fine-tuning baseline. These results suggest that \model is not merely a benchmark-specific artifact; it captures reward-hacking features that transfer beyond \benchmark and can generalize to other hacking patterns even in different domains such as code and math.

Overall, \model consistently improves robustness on targeted hacking categories while preserving general reward-model capability. 
Compared with data-based fine-tuning, the training-free vector-editing approach provides a stronger and more controllable robustness--capability trade-off.

\section{Mechanistic Insights}
\label{sec:major-insights}

\begin{table}[t]
\centering
\footnotesize
\setlength{\tabcolsep}{3pt}
\renewcommand{\arraystretch}{1.1}
\caption{
\textbf{Relationship between hacking directions, presentation/style direction, and token length.}
For each reward model, we report the mean cosine similarity between the extracted hacking direction $v_k$ and the presentation/style direction $v_{\mathrm{sty}}$, the mean hacked-minus-gold token-count difference $\overline{\Delta\mathrm{tok}}$, and the Pearson correlation between $v_{\mathrm{sty}}$ projection and token-count difference.
}
\label{tab:style-vs-length}
\resizebox{0.5\columnwidth}{!}{%
\begin{tabular}{@{}ll ccc@{}}
\toprule
\textbf{RM} & \textbf{Size}
 & \textbf{$\cos(v_k, v_{\mathrm{sty}})$}
 & \textbf{$\overline{\Delta\mathrm{tok}}$}
 & \textbf{Pearson $\rho$} \\
\midrule
Skywork-V2-Qwen3         & 0.6B & $+0.51$ & $+20.24$ & $+0.39$ \\
InternLM2-reward         & 1.8B & $+0.65$ & $-5.34$  & $+0.03$ \\
GRM-Llama-3.2            & 3B   & $+0.46$ & $+14.82$ & $+0.30$ \\
Skywork-V2-Llama-3.2     & 3B   & $+0.44$ & $+3.20$  & $+0.16$ \\
RM-Mistral               & 7B   & $+0.20$ & $+8.61$  & $+0.15$ \\
FsfairX-LLaMA3-RM        & 8B   & $+0.27$ & $+7.86$  & $+0.19$ \\
Skywork-Reward-Llama-3.1 & 8B   & $+0.52$ & $+8.32$  & $+0.30$ \\
InternLM2-reward         & 20B  & $+0.17$ & $+8.95$  & $+0.06$ \\
\midrule
\textbf{Mean across RMs} &      & $\mathbf{+0.40}$ & $\mathbf{+8.33}$ & $\mathbf{+0.20}$ \\
\bottomrule
\end{tabular}
}
\end{table}

\begin{table*}[h]
\centering
\scriptsize
\setlength{\tabcolsep}{2.5pt}
\renewcommand{\arraystretch}{1.15}
\caption{
\textbf{Component ablation analysis of hacking directions.}
For each reward model, we report the baseline gold-preference rate (Base, GP \%) and the change in gold-preference rate after ablating three directions: the shared presentation/style direction $v_{\mathrm{sty}}$, the style-orthogonal category-specific residual $v_{k,\perp\mathrm{sty}}$, and the full hacking direction $v_k$ (Joint).
Results are reported separately on target and non-target subcategories.
All ablation columns are changes relative to the baseline in percentage points (pp; $\uparrow$ means higher is better).
}
\label{tab:super-additivity}
\resizebox{0.9\textwidth}{!}{%
\begin{tabular}{ll c rrr c rrr}
\toprule
& & \multicolumn{4}{c}{\textbf{Target subcategories}} & \multicolumn{4}{c}{\textbf{Non-target subcategories}} \\
\cmidrule(lr){3-6}\cmidrule(lr){7-10}
\textbf{RM} & \textbf{Size}
 & \textbf{Base}
 & \textbf{$v_{\mathrm{sty}}$}
 & \textbf{$v_{k,\perp\mathrm{sty}}$}
 & \textbf{Joint}
 & \textbf{Base}
 & \textbf{$v_{\mathrm{sty}}$}
 & \textbf{$v_{k,\perp\mathrm{sty}}$}
 & \textbf{Joint} \\
 & & (GP \%) & (pp $\uparrow$) & (pp $\uparrow$) & (pp $\uparrow$)
 & (GP \%) & (pp $\uparrow$) & (pp $\uparrow$) & (pp $\uparrow$) \\
\midrule
Skywork-V2-Qwen3         & 0.6B & $52.16$ & $-3.99$  & $+5.95$ & $\mathbf{+27.89}$ & $87.82$ & $+2.85$  & $-0.01$ & $\mathbf{+1.76}$ \\
InternLM2-reward         & 1.8B & $58.93$ & $-15.36$ & $+3.93$ & $\mathbf{+19.64}$ & $89.66$ & $-11.16$ & $+1.50$ & $\mathbf{+1.45}$ \\
GRM-Llama-3.2            & 3B   & $45.70$ & $+8.58$  & $+2.80$ & $\mathbf{+27.26}$ & $85.24$ & $+3.96$  & $+2.87$ & $\mathbf{+3.95}$ \\
Skywork-V2-Llama-3.2     & 3B   & $73.89$ & $+1.09$  & $+6.93$ & $\mathbf{+10.91}$ & $89.88$ & $-0.01$  & $+1.53$ & $\mathbf{+3.00}$ \\
RM-Mistral               & 7B   & $67.97$ & $+3.52$  & $+5.14$ & $\mathbf{+13.00}$ & $93.96$ & $+0.78$  & $-0.04$ & $\mathbf{+2.58}$ \\
FsfairX-LLaMA3-RM        & 8B   & $58.21$ & $-2.05$  & $+7.66$ & $\mathbf{+13.77}$ & $89.26$ & $+3.48$  & $+0.54$ & $\mathbf{+2.08}$ \\
Skywork-Reward-Llama-3.1 & 8B   & $65.98$ & $+1.36$  & $+1.80$ & $\mathbf{+12.24}$ & $94.94$ & $+0.81$  & $-0.81$ & $\mathbf{+0.67}$ \\
InternLM2-reward         & 20B  & $73.26$ & $+0.36$  & $+8.34$ & $\mathbf{+8.68}$  & $92.45$ & $+0.36$  & $+0.36$ & $\mathbf{+1.07}$ \\
\midrule
\textbf{Mean}            &      & $\mathbf{62.01}$ & $\mathbf{-0.81}$ & $\mathbf{+5.32}$ & $\mathbf{+16.67}$ & $\mathbf{90.40}$ & $\mathbf{+0.13}$ & $\mathbf{+0.74}$ & $\mathbf{+2.07}$ \\
\bottomrule
\end{tabular}
}
\end{table*}

\subsection{Diagnostic Style--Category Decomposition}
\label{sec:carve-decomposition}

We further use the extracted hacking directions to diagnose whether each direction mainly reflects general presentation-related features, such as length, polish, and confidence, or also contains category-specific hacking signals. For each reward model, we estimate a unit \emph{style direction} $v_{\mathrm{sty}}$ from non-target gold responses by contrasting the final hidden states of responses in the top and bottom token-count quartiles, which captures presentation features such as length, elaboration, vocabulary complexity, and confidence.
We then decompose each hacking direction $v_k$ into a style-aligned component, which captures overlap with shared presentation features, and a style-orthogonal residual, which captures the remaining category-specific signal.
\begin{equation}
v_k =
\underbrace{(v_k^\top v_{\mathrm{sty}}) v_{\mathrm{sty}}}_{\text{style-aligned}}
+
\underbrace{v_{k,\perp \mathrm{sty}}}_{\text{style-orthogonal residual}},
\label{eq:carve-decomp}
\end{equation}
where $v_{k,\perp \mathrm{sty}}$ is orthogonal to $v_{\mathrm{sty}}$. 
We run three diagnostic experiments with the decomposition: removing only the shared style direction $v_{\mathrm{sty}}$, removing only the style-orthogonal residual directions $v_{k,\perp \mathrm{sty}}$, and removing the full hacking directions.

\subsection{Results and Insights}

\paragraph{The style direction captures style and elaboration, not literal length.}

We audit the relationship between extracted hacking directions, the reward head, and the style direction using cosine similarity. \Cref{tab:style-vs-length} shows that the extracted directions are substantially entangled with the style direction, with mean $\cos(\hat{v}_k, v_{\mathrm{sty}})=+0.40$. 
However, this style component is not simply token length, as the mean correlation between $v_{\mathrm{sty}}$ projection and token-count difference is only $+0.20$, which suggests that \model{} is not merely removing a length feature, but a broader stylistic cluster that reward models associate with hacked responses.

\paragraph{Joint editing is super-additive.}

\Cref{tab:super-additivity} shows that the full hacking direction is more effective than the decomposed style or style-orthogonal hacking category-related components. On target subcategories, ablating only the style direction or only the style-orthogonal residual gives limited gains, while ablating the raw joint direction improves gold-preference rate by $+16.67$ percentage points on average. This indicates that reward hacking is not captured by an isolated style feature or an isolated category-specific feature. Instead, reward hacking arises from jointly interacting residual-stream directions, rather than from independent single-feature biases.
On non-target subcategories, reward models' performance is already high, and joint ablation changes gold-preference rate by only $+2.1$ points on average. This supports the interpretation that \model{} primarily removes target-relevant hacking sensitivity without globally degrading the reward model.

\section{Ablation Studies}
\label{sec:ablation}

\subsection{Fooled-Subset Filtering}
\label{sec:ablation-fooled}

\begin{table}[t]
\centering
\small
\setlength{\tabcolsep}{3pt}
\renewcommand{\arraystretch}{1.1}
\caption{
\textbf{Effect of fooled-subset filtering on hacking-direction estimation.}
\textsc{All} reports the cosine alignment between the reward-head direction $w_r$ and the hacked-minus-gold hidden-state difference when the direction is estimated from all training pairs; \textsc{Fooled} reports the corresponding cosine alignment when the direction is estimated only from fooled pairs.
Values are averaged over each reward model's target subcategories.
$\Delta=\textsc{Fooled}-\textsc{All}$.
}
\label{tab:phase1-fooled-filter}
\resizebox{0.5\linewidth}{!}{%
\begin{tabular}{l@{\hspace{6pt}}l@{\hspace{6pt}}rrr}
\toprule
\textbf{RM} & \textbf{Size}
  & \textbf{All} & \textbf{Fooled} & \textbf{$\Delta$} \\
\midrule
Skywork-V2-Qwen3       & 0.6B & $+0.06$ & $+0.17$ & $+0.12$ \\
InternLM2-reward       & 1.8B & $+0.12$ & $+0.52$ & $+0.39$ \\
GRM-Llama-3.2          & 3B   & $+0.03$ & $+0.23$ & $+0.20$ \\
Skywork-V2-Llama-3.2   & 3B   & $-0.21$ & $+0.20$ & $+0.41$ \\
RM-Mistral             & 7B   & $-0.05$ & $+0.10$ & $+0.14$ \\
FsfairX-LLaMA3-RM      & 8B   & $\phantom{-}0.00$ & $+0.11$ & $+0.11$ \\
Skywork-Reward-Llama-3.1 & 8B & $-0.09$ & $+0.29$ & $+0.38$ \\
InternLM2-reward       & 20B  & $-0.28$ & $+0.25$ & $+0.53$ \\
\midrule
\textbf{Mean}          &      & $\mathbf{-0.05}$ & $\mathbf{+0.24}$ & $\mathbf{+0.29}$ \\
\bottomrule
\end{tabular}
}
\end{table}

\begin{figure*}[!htb]
    \centering
    \includegraphics[width=0.99\textwidth]{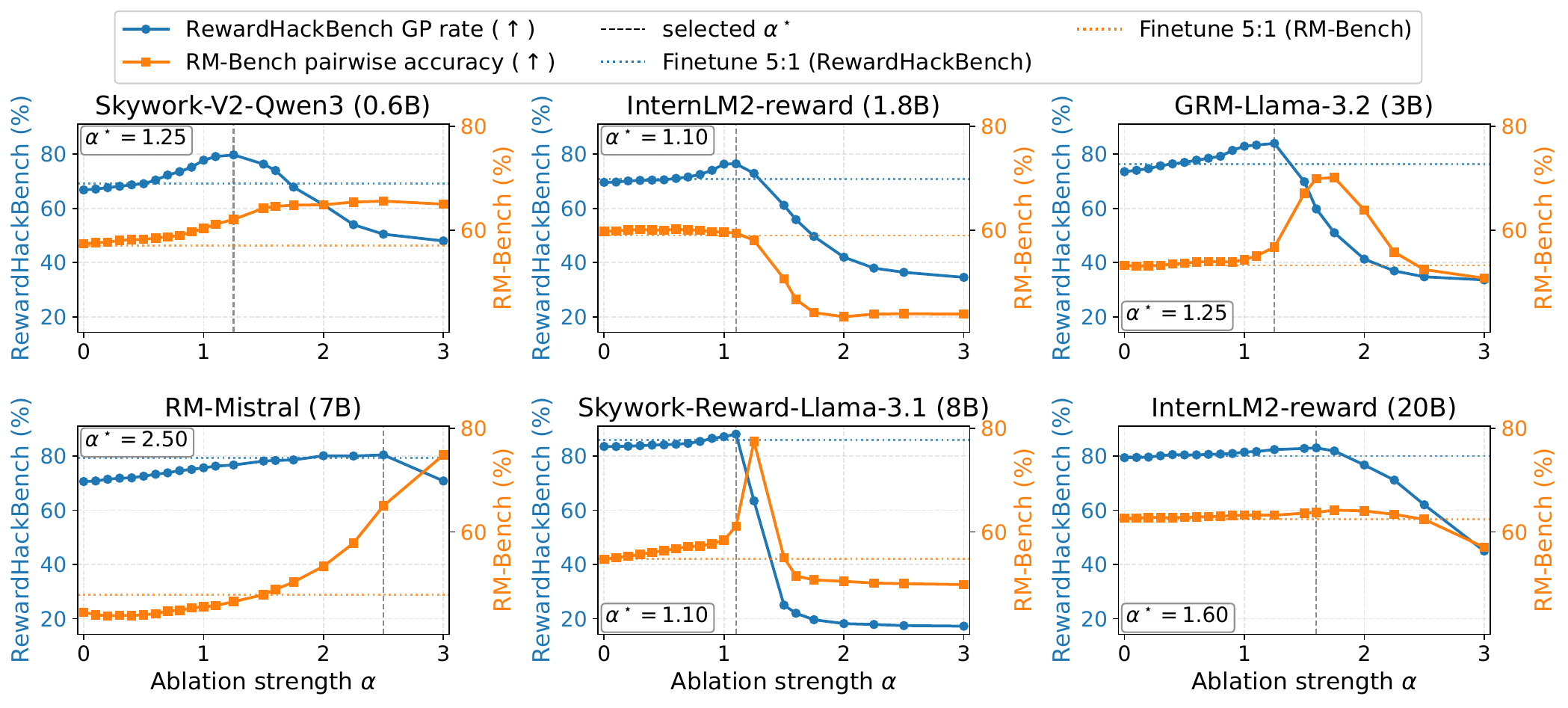}
    \caption{\textbf{Trade-off curves over intervention strength $\alpha$ for six representative reward models.} 
    Blue curves show \benchmark gold-preference rate (GP rate), and orange curves show RM-Bench Hard accuracy. 
    Dashed vertical lines mark the selected $\alpha^\star$ for each RM, and dotted horizontal lines show the 5:1 fine-tuning baseline. 
    The full eight-model figure is provided in Appendix~\ref{app: tradeoff}.
    }
    \label{fig:alpha-tradeoff-main}
\end{figure*}

For each reward model, we compare the cosine alignment between the reward-head direction $w_r$ and the hacked-minus-gold hidden-state difference estimated in two ways: using all training pairs, and using only fooled pairs where the base reward model assigns a higher score to the hacked response than to the gold response. 
As shown in \Cref{tab:phase1-fooled-filter}, fooled-subset filtering increases the cosine alignment from $-0.05$ with all pairs to $+0.24$. 
This indicates that fooled examples better isolate the hidden-state direction that the reward model actually rewards when it prefers hacked responses.

\subsection{Intervention Strength $\alpha$}
\label{sec:ablation-alpha}

We further conduct ablation studies to examine how the editing strength $\alpha$ controls the robustness--capability trade-off. For each reward model, we sweep $\alpha$ from $0$ to $3$ and track performance on both \benchmark{} and RM-Bench. As shown in \Cref{fig:alpha-tradeoff-main}, increasing $\alpha$ generally improves \benchmark performance at first, but overly large values can eventually degrade performance on either \benchmark or RM-Bench.  This motivates selecting a per-model sweet spot $\alpha^\star$ rather than using a fixed global intervention strength. In our main experiments, $\alpha^\star$ is chosen to maximize \benchmark improvement while preserving RM-Bench performance.

\section{Related Work}

Existing reward hacking mitigation methods include data-centric augmentation or rebalancing \citep{shen2023trickle, liu2025rrm}, training-based changes to reward-model parameters \citep{chen2024odin, liu2025rrm}, and optimization-time methods such as reward ensembles \citep{Coste2024, eisenstein2023helping}. 
These approaches can reduce reward hacking, but often require additional curated data, costly reward-model updates, or inference-time aggregation. In contrast, \model{} edits only the reward-head vector, requiring no gradient updates and leaving the transformer backbone unchanged. Additional related work is provided in Appendix~\ref{appendix-related-work}.

\section{Conclusion}
\label{sec:conclusion}

We introduce \benchmark{}, a benchmark covering professional-domain and general-purpose reward-hacking patterns, and show that eight open-source scalar reward models remain vulnerable across model sizes and families. 
Motivated by these failures, we present \model, a training-free reward-head editing method for mitigating reward hacking. 
Across eight reward models, \model consistently outperforms fine-tuning baselines on target hacking subcategories and largely preserves broader reward-model capability, including performance on RM-Bench. 
Further analyses show that removing the full multi-directional hacking subspace is more effective than removing its style-related or category-specific components separately, suggesting that these failures arise from jointly interacting residual-stream directions.

\clearpage

\section*{Limitations}
This work has several limitations. 
First, \model applies to scalar reward models with accessible reward-head vectors and hidden states, and does not directly cover closed-source reward models or generative pairwise evaluators. 
Second, although \benchmark includes both professional-domain and general-purpose reward-hacking patterns, it cannot exhaust all real-world failure modes, especially in high-stakes domains where errors may be more context-dependent. 
Third, \model{} requires a small set of contrastive gold--hacked examples and selects intervention strength using validation trade-off curves, so its effectiveness may depend on the coverage of these examples. 
Finally, we evaluate reward-model robustness directly, but do not study downstream effects when the edited reward models are used inside full RLHF or inference-time optimization pipelines.

\bibliographystyle{IEEEtranN}
\bibliography{main}

\begin{thebibliography}{31}
\providecommand{\natexlab}[1]{#1}
\providecommand{\url}[1]{#1}
\csname url@samestyle\endcsname
\providecommand{\newblock}{\relax}
\providecommand{\bibinfo}[2]{#2}
\providecommand{\BIBentrySTDinterwordspacing}{\spaceskip=0pt\relax}
\providecommand{\BIBentryALTinterwordstretchfactor}{4}
\providecommand{\BIBentryALTinterwordspacing}{\spaceskip=\fontdimen2\font plus
\BIBentryALTinterwordstretchfactor\fontdimen3\font minus \fontdimen4\font\relax}
\providecommand{\BIBforeignlanguage}[2]{{%
\expandafter\ifx\csname l@#1\endcsname\relax
\typeout{** WARNING: IEEEtranN.bst: No hyphenation pattern has been}%
\typeout{** loaded for the language `#1'. Using the pattern for}%
\typeout{** the default language instead.}%
\else
\language=\csname l@#1\endcsname
\fi
#2}}
\providecommand{\BIBdecl}{\relax}
\BIBdecl

\bibitem[Stiennon et~al.(2020)Stiennon, Ouyang, Wu, Ziegler, Lowe, Voss, Radford, Amodei, and Christiano]{Stiennon2020}
\BIBentryALTinterwordspacing
N.~Stiennon, L.~Ouyang, J.~Wu, D.~Ziegler, R.~Lowe, C.~Voss, A.~Radford, D.~Amodei, and P.~F. Christiano, ``Learning to summarize with human feedback,'' in \emph{Advances in Neural Information Processing Systems}, H.~Larochelle, M.~Ranzato, R.~Hadsell, M.~Balcan, and H.~Lin, Eds., vol.~33.\hskip 1em plus 0.5em minus 0.4em\relax Curran Associates, Inc., 2020, pp. 3008--3021. [Online]. Available: \url{https://proceedings.neurips.cc/paper_files/paper/2020/file/1f89885d556929e98d3ef9b86448f951-Paper.pdf}
\BIBentrySTDinterwordspacing

\bibitem[Ouyang et~al.(2022)Ouyang, Wu, Jiang, Almeida, Wainwright, Mishkin, Zhang, Agarwal, Slama, Ray, Schulman, Hilton, Kelton, Miller, Simens, Askell, Welinder, Christiano, Leike, and Lowe]{ouyang2022training}
\BIBentryALTinterwordspacing
L.~Ouyang, J.~Wu, X.~Jiang, D.~Almeida, C.~Wainwright, P.~Mishkin, C.~Zhang, S.~Agarwal, K.~Slama, A.~Ray, J.~Schulman, J.~Hilton, F.~Kelton, L.~Miller, M.~Simens, A.~Askell, P.~Welinder, P.~F. Christiano, J.~Leike, and R.~Lowe, ``Training language models to follow instructions with human feedback,'' in \emph{Advances in Neural Information Processing Systems}, S.~Koyejo, S.~Mohamed, A.~Agarwal, D.~Belgrave, K.~Cho, and A.~Oh, Eds., vol.~35.\hskip 1em plus 0.5em minus 0.4em\relax Curran Associates, Inc., 2022, pp. 27\,730--27\,744. [Online]. Available: \url{https://proceedings.neurips.cc/paper_files/paper/2022/file/b1efde53be364a73914f58805a001731-Paper-Conference.pdf}
\BIBentrySTDinterwordspacing

\bibitem[Singhal et~al.(2024)Singhal, Goyal, Xu, and Durrett]{singhal2023long}
\BIBentryALTinterwordspacing
P.~Singhal, T.~Goyal, J.~Xu, and G.~Durrett, ``A long way to go: Investigating length correlations in rlhf,'' 2024. [Online]. Available: \url{https://arxiv.org/abs/2310.03716}
\BIBentrySTDinterwordspacing

\bibitem[Gao et~al.(2023)Gao, Schulman, and Hilton]{gao2023scaling}
\BIBentryALTinterwordspacing
L.~Gao, J.~Schulman, and J.~Hilton, ``Scaling laws for reward model overoptimization,'' in \emph{Proceedings of the 40th International Conference on Machine Learning}, ser. Proceedings of Machine Learning Research, A.~Krause, E.~Brunskill, K.~Cho, B.~Engelhardt, S.~Sabato, and J.~Scarlett, Eds., vol. 202.\hskip 1em plus 0.5em minus 0.4em\relax PMLR, 23--29 Jul 2023, pp. 10\,835--10\,866. [Online]. Available: \url{https://proceedings.mlr.press/v202/gao23h.html}
\BIBentrySTDinterwordspacing

\bibitem[Park et~al.(2024)Park, Jwa, Meiying, Kim, and Choi]{park2024offsetbias}
\BIBentryALTinterwordspacing
J.~Park, S.~Jwa, R.~Meiying, D.~Kim, and S.~Choi, ``{O}ffset{B}ias: Leveraging debiased data for tuning evaluators,'' in \emph{Findings of the Association for Computational Linguistics: EMNLP 2024}, Y.~Al-Onaizan, M.~Bansal, and Y.-N. Chen, Eds.\hskip 1em plus 0.5em minus 0.4em\relax Miami, Florida, USA: Association for Computational Linguistics, Nov. 2024, pp. 1043--1067. [Online]. Available: \url{https://aclanthology.org/2024.findings-emnlp.57/}
\BIBentrySTDinterwordspacing

\bibitem[Liu et~al.(2025{\natexlab{a}})Liu, Yao, Min, Cao, Hou, and Li]{liu2025rm}
\BIBentryALTinterwordspacing
Y.~Liu, Z.~Yao, R.~Min, Y.~Cao, L.~Hou, and J.~Li, ``Rm-bench: Benchmarking reward models of language models with subtlety and style,'' in \emph{International Conference on Learning Representations}, Y.~Yue, A.~Garg, N.~Peng, F.~Sha, and R.~Yu, Eds., vol. 2025, 2025, pp. 44\,323--44\,355. [Online]. Available: \url{https://proceedings.iclr.cc/paper_files/paper/2025/file/6da1eec80095dc5937f7716db15aca4b-Paper-Conference.pdf}
\BIBentrySTDinterwordspacing

\bibitem[Sharma et~al.(2024)Sharma, Tong, Korbak, Duvenaud, Askell, Bowman, DURMUS, Hatfield-Dodds, Johnston, Kravec, Maxwell, McCandlish, Ndousse, Rausch, Schiefer, Yan, Zhang, and Perez]{sharma2024sycophancy}
\BIBentryALTinterwordspacing
M.~Sharma, M.~Tong, T.~Korbak, D.~Duvenaud, A.~Askell, S.~Bowman, E.~DURMUS, Z.~Hatfield-Dodds, S.~Johnston, S.~Kravec, T.~Maxwell, S.~McCandlish, K.~Ndousse, O.~Rausch, N.~Schiefer, D.~Yan, M.~Zhang, and E.~Perez, ``Towards understanding sycophancy in language models,'' in \emph{International Conference on Learning Representations}, B.~Kim, Y.~Yue, S.~Chaudhuri, K.~Fragkiadaki, M.~Khan, and Y.~Sun, Eds., vol. 2024, 2024, pp. 110--144. [Online]. Available: \url{https://proceedings.iclr.cc/paper_files/paper/2024/file/0105f7972202c1d4fb817da9f21a9663-Paper-Conference.pdf}
\BIBentrySTDinterwordspacing

\bibitem[Lambert et~al.(2025)Lambert, Pyatkin, Morrison, Miranda, Lin, Chandu, Dziri, Kumar, Zick, Choi, Smith, and Hajishirzi]{lambert-etal-2025-rewardbench}
\BIBentryALTinterwordspacing
N.~Lambert, V.~Pyatkin, J.~Morrison, L.~Miranda, B.~Y. Lin, K.~Chandu, N.~Dziri, S.~Kumar, T.~Zick, Y.~Choi, N.~A. Smith, and H.~Hajishirzi, ``{R}eward{B}ench: Evaluating reward models for language modeling,'' in \emph{Findings of the Association for Computational Linguistics: NAACL 2025}, L.~Chiruzzo, A.~Ritter, and L.~Wang, Eds.\hskip 1em plus 0.5em minus 0.4em\relax Albuquerque, New Mexico: Association for Computational Linguistics, Apr. 2025, pp. 1755--1797. [Online]. Available: \url{https://aclanthology.org/2025.findings-naacl.96/}
\BIBentrySTDinterwordspacing

\bibitem[Shen et~al.(2024)Shen, Chen, Song, Jin, Peng, Mi, Khashabi, and Yu]{shen2023trickle}
\BIBentryALTinterwordspacing
L.~Shen, S.~Chen, L.~Song, L.~Jin, B.~Peng, H.~Mi, D.~Khashabi, and D.~Yu, ``The trickle-down impact of reward inconsistency on rlhf,'' in \emph{International Conference on Learning Representations}, B.~Kim, Y.~Yue, S.~Chaudhuri, K.~Fragkiadaki, M.~Khan, and Y.~Sun, Eds., vol. 2024, 2024, pp. 33\,029--33\,057. [Online]. Available: \url{https://proceedings.iclr.cc/paper_files/paper/2024/file/8c976a95df6a229551cd28c76627edc9-Paper-Conference.pdf}
\BIBentrySTDinterwordspacing

\bibitem[Liu et~al.(2025{\natexlab{b}})Liu, Xiong, Ren, Chen, Wu, Joshi, Gao, Shen, Qin, Yu, Sohn, Makarova, Liu, Liu, Piot, Ittycheriah, Kumar, and Saleh]{liu2025rrm}
\BIBentryALTinterwordspacing
T.~Liu, W.~Xiong, J.~Ren, L.~Chen, J.~Wu, R.~Joshi, Y.~Gao, J.~Shen, Z.~Qin, T.~Yu, D.~Sohn, A.~Makarova, J.~Z. Liu, Y.~Liu, B.~Piot, A.~Ittycheriah, A.~Kumar, and M.~Saleh, ``Rrm: Robust reward model training mitigates reward hacking,'' in \emph{International Conference on Learning Representations}, Y.~Yue, A.~Garg, N.~Peng, F.~Sha, and R.~Yu, Eds., vol. 2025, 2025, pp. 62\,682--62\,700. [Online]. Available: \url{https://proceedings.iclr.cc/paper_files/paper/2025/file/9d46574e5baae5121180228223a11836-Paper-Conference.pdf}
\BIBentrySTDinterwordspacing

\bibitem[Chen et~al.(2024)Chen, Zhu, Chen, Soselia, Zhou, Goldstein, Huang, Shoeybi, and Catanzaro]{chen2024odin}
\BIBentryALTinterwordspacing
L.~Chen, C.~Zhu, J.~Chen, D.~Soselia, T.~Zhou, T.~Goldstein, H.~Huang, M.~Shoeybi, and B.~Catanzaro, ``{ODIN}: Disentangled reward mitigates hacking in {RLHF},'' in \emph{Proceedings of the 41st International Conference on Machine Learning}, ser. Proceedings of Machine Learning Research, R.~Salakhutdinov, Z.~Kolter, K.~Heller, A.~Weller, N.~Oliver, J.~Scarlett, and F.~Berkenkamp, Eds., vol. 235.\hskip 1em plus 0.5em minus 0.4em\relax PMLR, 21--27 Jul 2024, pp. 7935--7952. [Online]. Available: \url{https://proceedings.mlr.press/v235/chen24bn.html}
\BIBentrySTDinterwordspacing

\bibitem[Zeng et~al.(2024)Zeng, Yu, Gao, Meng, Goyal, and Chen]{zeng2024llmbar}
\BIBentryALTinterwordspacing
Z.~Zeng, J.~Yu, T.~Gao, Y.~Meng, T.~Goyal, and D.~Chen, ``Evaluating large language models at evaluating instruction following,'' in \emph{International Conference on Learning Representations}, B.~Kim, Y.~Yue, S.~Chaudhuri, K.~Fragkiadaki, M.~Khan, and Y.~Sun, Eds., vol. 2024, 2024, pp. 40\,193--40\,219. [Online]. Available: \url{https://proceedings.iclr.cc/paper_files/paper/2024/file/afc8b034823271816d14f7c1aefe1dff-Paper-Conference.pdf}
\BIBentrySTDinterwordspacing

\bibitem[Guha et~al.(2023)Guha, Nyarko, Ho, R\'{e}, Chilton, K, Chohlas-Wood, Peters, Waldon, Rockmore, Zambrano, Talisman, Hoque, Surani, Fagan, Sarfaty, Dickinson, Porat, Hegland, Wu, Nudell, Niklaus, Nay, Choi, Tobia, Hagan, Ma, Livermore, Rasumov-Rahe, Holzenberger, Kolt, Henderson, Rehaag, Goel, Gao, Williams, Gandhi, Zur, Iyer, and Li]{guha2023legalbench}
\BIBentryALTinterwordspacing
N.~Guha, J.~Nyarko, D.~Ho, C.~R\'{e}, A.~Chilton, A.~K, A.~Chohlas-Wood, A.~Peters, B.~Waldon, D.~Rockmore, D.~Zambrano, D.~Talisman, E.~Hoque, F.~Surani, F.~Fagan, G.~Sarfaty, G.~Dickinson, H.~Porat, J.~Hegland, J.~Wu, J.~Nudell, J.~Niklaus, J.~Nay, J.~Choi, K.~Tobia, M.~Hagan, M.~Ma, M.~Livermore, N.~Rasumov-Rahe, N.~Holzenberger, N.~Kolt, P.~Henderson, S.~Rehaag, S.~Goel, S.~Gao, S.~Williams, S.~Gandhi, T.~Zur, V.~Iyer, and Z.~Li, ``Legalbench: A collaboratively built benchmark for measuring legal reasoning in large language models,'' in \emph{Advances in Neural Information Processing Systems}, A.~Oh, T.~Naumann, A.~Globerson, K.~Saenko, M.~Hardt, and S.~Levine, Eds., vol.~36.\hskip 1em plus 0.5em minus 0.4em\relax Curran Associates, Inc., 2023, pp. 44\,123--44\,279. [Online]. Available: \url{https://proceedings.neurips.cc/paper_files/paper/2023/file/89e44582fd28ddfea1ea4dcb0ebbf4b0-Paper-Datasets_and_Benchmarks.pdf}
\BIBentrySTDinterwordspacing

\bibitem[Fan et~al.(2026)Fan, Ni, Merane, Tian, Hermstrüwer, Huang, Akhtar, Salimbeni, Geering, Dreyer, Brunner, Leippold, Sachan, Stremitzer, Engel, Ash, and Niklaus]{lexam2024}
\BIBentryALTinterwordspacing
Y.~Fan, J.~Ni, J.~Merane, Y.~Tian, Y.~Hermstrüwer, Y.~Huang, M.~Akhtar, E.~Salimbeni, F.~Geering, O.~Dreyer, D.~Brunner, M.~Leippold, M.~Sachan, A.~Stremitzer, C.~Engel, E.~Ash, and J.~Niklaus, ``Lexam: Benchmarking legal reasoning on 340 law exams,'' 2026. [Online]. Available: \url{https://arxiv.org/abs/2505.12864}
\BIBentrySTDinterwordspacing

\bibitem[Zheng et~al.(2025)Zheng, Guha, Arifov, Zhang, Skreta, Manning, Henderson, and Ho]{zheng2025barexam}
\BIBentryALTinterwordspacing
L.~Zheng, N.~Guha, J.~Arifov, S.~Zhang, M.~Skreta, C.~D. Manning, P.~Henderson, and D.~E. Ho, ``A reasoning-focused legal retrieval benchmark,'' in \emph{Proceedings of the 2025 Symposium on Computer Science and Law}, ser. CSLAW '25.\hskip 1em plus 0.5em minus 0.4em\relax New York, NY, USA: Association for Computing Machinery, 2025, p. 169–193. [Online]. Available: \url{https://doi.org/10.1145/3709025.3712219}
\BIBentrySTDinterwordspacing

\bibitem[Liu et~al.(2026{\natexlab{a}})Liu, Zeng, Xiao, He, Liu, Wang, Yan, Shen, Zhang, Xu, Liu, and Zhou]{liu2025skywork}
\BIBentryALTinterwordspacing
C.~Y. Liu, L.~Zeng, Y.~Xiao, J.~He, J.~Liu, C.~Wang, R.~Yan, W.~Shen, F.~Zhang, J.~Xu, Y.~Liu, and Y.~Zhou, ``Skywork-reward-v2: Scaling preference data curation via human-ai synergy,'' 2026. [Online]. Available: \url{https://arxiv.org/abs/2507.01352}
\BIBentrySTDinterwordspacing

\bibitem[Cai et~al.(2024)Cai, Cao, Chen, Chen, Chen, Chen, Chen, Chen, Chen, Chu, Dong, Duan, Fan, Fei, Gao, Ge, Gu, Gu, Gui, Guo, Guo, He, Hu, Huang, Jiang, Jiao, Jin, Lei, Li, Li, Li, Li, Li, Li, Liu, Liu, Hong, Liu, Liu, Liu, Lv, Lv, Lv, Ma, Ma, Ma, Ning, Ouyang, Qiu, Qu, Shang, Shao, Song, Song, Sui, Sun, Sun, Tang, Wang, Wang, Wang, Wang, Wang, Wang, Wang, Wei, Weng, Wu, Xiong, Xu, Xu, Yan, Yan, Yang, Ye, Ying, Yu, Yu, Zang, Zhang, Zhang, Zhang, Zhang, Zhang, Zhang, Zhang, Zhang, Zhang, Zhang, Zhang, Zhao, Zhao, Zhao, Zhou, Zhou, Zhuo, Zou, Qiu, Qiao, and Lin]{cai2024internlm2}
\BIBentryALTinterwordspacing
Z.~Cai, M.~Cao, H.~Chen, K.~Chen, K.~Chen, X.~Chen, X.~Chen, Z.~Chen, Z.~Chen, P.~Chu, X.~Dong, H.~Duan, Q.~Fan, Z.~Fei, Y.~Gao, J.~Ge, C.~Gu, Y.~Gu, T.~Gui, A.~Guo, Q.~Guo, C.~He, Y.~Hu, T.~Huang, T.~Jiang, P.~Jiao, Z.~Jin, Z.~Lei, J.~Li, J.~Li, L.~Li, S.~Li, W.~Li, Y.~Li, H.~Liu, J.~Liu, J.~Hong, K.~Liu, K.~Liu, X.~Liu, C.~Lv, H.~Lv, K.~Lv, L.~Ma, R.~Ma, Z.~Ma, W.~Ning, L.~Ouyang, J.~Qiu, Y.~Qu, F.~Shang, Y.~Shao, D.~Song, Z.~Song, Z.~Sui, P.~Sun, Y.~Sun, H.~Tang, B.~Wang, G.~Wang, J.~Wang, J.~Wang, R.~Wang, Y.~Wang, Z.~Wang, X.~Wei, Q.~Weng, F.~Wu, Y.~Xiong, C.~Xu, R.~Xu, H.~Yan, Y.~Yan, X.~Yang, H.~Ye, H.~Ying, J.~Yu, J.~Yu, Y.~Zang, C.~Zhang, L.~Zhang, P.~Zhang, P.~Zhang, R.~Zhang, S.~Zhang, S.~Zhang, W.~Zhang, W.~Zhang, X.~Zhang, X.~Zhang, H.~Zhao, Q.~Zhao, X.~Zhao, F.~Zhou, Z.~Zhou, J.~Zhuo, Y.~Zou, X.~Qiu, Y.~Qiao, and D.~Lin, ``Internlm2 technical report,'' 2024. [Online]. Available: \url{https://arxiv.org/abs/2403.17297}
\BIBentrySTDinterwordspacing

\bibitem[Yang et~al.(2024)Yang, Ding, Lin, Zhang, and Zhang]{yang2024GRM}
\BIBentryALTinterwordspacing
R.~Yang, R.~Ding, Y.~Lin, H.~Zhang, and T.~Zhang, ``Regularizing hidden states enables learning generalizable reward model for llms,'' 2024. [Online]. Available: \url{https://arxiv.org/abs/2406.10216}
\BIBentrySTDinterwordspacing

\bibitem[Dong et~al.(2023)Dong, Xiong, Goyal, Zhang, Chow, Pan, Diao, Zhang, Shum, and Zhang]{dong2023raft}
\BIBentryALTinterwordspacing
H.~Dong, W.~Xiong, D.~Goyal, Y.~Zhang, W.~Chow, R.~Pan, S.~Diao, J.~Zhang, K.~Shum, and T.~Zhang, ``Raft: Reward ranked finetuning for generative foundation model alignment,'' 2023. [Online]. Available: \url{https://arxiv.org/abs/2304.06767}
\BIBentrySTDinterwordspacing

\bibitem[Xiong et~al.(2024)Xiong, Dong, Ye, Wang, Zhong, Ji, Jiang, and Zhang]{xiong2024iterative}
\BIBentryALTinterwordspacing
W.~Xiong, H.~Dong, C.~Ye, Z.~Wang, H.~Zhong, H.~Ji, N.~Jiang, and T.~Zhang, ``Iterative preference learning from human feedback: Bridging theory and practice for rlhf under kl-constraint,'' 2024. [Online]. Available: \url{https://arxiv.org/abs/2312.11456}
\BIBentrySTDinterwordspacing

\bibitem[Dong et~al.(2024)Dong, Xiong, Pang, Wang, Zhao, Zhou, Jiang, Sahoo, Xiong, and Zhang]{dong2024rlhf}
\BIBentryALTinterwordspacing
H.~Dong, W.~Xiong, B.~Pang, H.~Wang, H.~Zhao, Y.~Zhou, N.~Jiang, D.~Sahoo, C.~Xiong, and T.~Zhang, ``Rlhf workflow: From reward modeling to online rlhf,'' 2024. [Online]. Available: \url{https://arxiv.org/abs/2405.07863}
\BIBentrySTDinterwordspacing

\bibitem[Coste et~al.(2024)Coste, Anwar, Kirk, and Krueger]{Coste2024}
\BIBentryALTinterwordspacing
T.~Coste, U.~Anwar, R.~Kirk, and D.~Krueger, ``Reward model ensembles help mitigate overoptimization,'' in \emph{International Conference on Learning Representations}, B.~Kim, Y.~Yue, S.~Chaudhuri, K.~Fragkiadaki, M.~Khan, and Y.~Sun, Eds., vol. 2024, 2024, pp. 50\,905--50\,931. [Online]. Available: \url{https://proceedings.iclr.cc/paper_files/paper/2024/file/dda7f9378a210c25e470e19304cce85d-Paper-Conference.pdf}
\BIBentrySTDinterwordspacing

\bibitem[Eisenstein et~al.(2024)Eisenstein, Nagpal, Agarwal, Beirami, D'Amour, Dvijotham, Fisch, Heller, Pfohl, Ramachandran, Shaw, and Berant]{eisenstein2023helping}
\BIBentryALTinterwordspacing
J.~Eisenstein, C.~Nagpal, A.~Agarwal, A.~Beirami, A.~D'Amour, D.~Dvijotham, A.~Fisch, K.~Heller, S.~Pfohl, D.~Ramachandran, P.~Shaw, and J.~Berant, ``Helping or herding? reward model ensembles mitigate but do not eliminate reward hacking,'' 2024. [Online]. Available: \url{https://arxiv.org/abs/2312.09244}
\BIBentrySTDinterwordspacing

\bibitem[Skalse et~al.(2022)Skalse, Howe, Krasheninnikov, and Krueger]{Skalse2022}
\BIBentryALTinterwordspacing
J.~Skalse, N.~Howe, D.~Krasheninnikov, and D.~Krueger, ``Defining and characterizing reward gaming,'' in \emph{Advances in Neural Information Processing Systems}, S.~Koyejo, S.~Mohamed, A.~Agarwal, D.~Belgrave, K.~Cho, and A.~Oh, Eds., vol.~35.\hskip 1em plus 0.5em minus 0.4em\relax Curran Associates, Inc., 2022, pp. 9460--9471. [Online]. Available: \url{https://proceedings.neurips.cc/paper_files/paper/2022/file/3d719fee332caa23d5038b8a90e81796-Paper-Conference.pdf}
\BIBentrySTDinterwordspacing

\bibitem[Frick et~al.(2025)Frick, Li, Chen, Chiang, Angelopoulos, Jiao, Zhu, Gonzalez, and Stoica]{Frick2025}
\BIBentryALTinterwordspacing
E.~Frick, T.~Li, C.~Chen, W.-L. Chiang, A.~Angelopoulos, J.~Jiao, B.~Zhu, J.~E. Gonzalez, and I.~Stoica, ``How to evaluate reward models for rlhf,'' in \emph{International Conference on Learning Representations}, Y.~Yue, A.~Garg, N.~Peng, F.~Sha, and R.~Yu, Eds., vol. 2025, 2025, pp. 18\,128--18\,163. [Online]. Available: \url{https://proceedings.iclr.cc/paper_files/paper/2025/file/2e01083b381b4865919b4915ef32e3d2-Paper-Conference.pdf}
\BIBentrySTDinterwordspacing

\bibitem[Liu et~al.(2025{\natexlab{c}})Liu, Li, Ma, Zhao, and Du]{liu2025contracteval}
\BIBentryALTinterwordspacing
S.~Liu, Z.~Li, R.~Ma, H.~Zhao, and M.~Du, ``Contracteval: Benchmarking llms for clause-level legal risk identification in commercial contracts,'' 2025. [Online]. Available: \url{https://arxiv.org/abs/2508.03080}
\BIBentrySTDinterwordspacing

\bibitem[Liu et~al.(2026{\natexlab{b}})Liu, Zhang, Ma, Deng, Zhu, Li, Li, Shen, and Du]{liu2026agentslaw}
\BIBentryALTinterwordspacing
S.~Liu, R.~Zhang, R.~Ma, Y.~Deng, L.~Zhu, J.~Li, Z.~Li, Z.~Shen, and M.~Du, ``Llm agents in law: Taxonomy, applications, and challenges,'' 2026. [Online]. Available: \url{https://arxiv.org/abs/2601.06216}
\BIBentrySTDinterwordspacing

\bibitem[Zou et~al.(2025)Zou, Phan, Chen, Campbell, Guo, Ren, Pan, Yin, Mazeika, Dombrowski, Goel, Li, Byun, Wang, Mallen, Basart, Koyejo, Song, Fredrikson, Kolter, and Hendrycks]{zou2023representation}
\BIBentryALTinterwordspacing
A.~Zou, L.~Phan, S.~Chen, J.~Campbell, P.~Guo, R.~Ren, A.~Pan, X.~Yin, M.~Mazeika, A.-K. Dombrowski, S.~Goel, N.~Li, M.~J. Byun, Z.~Wang, A.~Mallen, S.~Basart, S.~Koyejo, D.~Song, M.~Fredrikson, J.~Z. Kolter, and D.~Hendrycks, ``Representation engineering: A top-down approach to ai transparency,'' 2025. [Online]. Available: \url{https://arxiv.org/abs/2310.01405}
\BIBentrySTDinterwordspacing

\bibitem[Turner et~al.(2024)Turner, Thiergart, Leech, Udell, Vazquez, Mini, and MacDiarmid]{turner2023steering}
\BIBentryALTinterwordspacing
A.~M. Turner, L.~Thiergart, G.~Leech, D.~Udell, J.~J. Vazquez, U.~Mini, and M.~MacDiarmid, ``Steering language models with activation engineering,'' 2024. [Online]. Available: \url{https://arxiv.org/abs/2308.10248}
\BIBentrySTDinterwordspacing

\bibitem[Arditi et~al.(2024)Arditi, Obeso, Syed, Paleka, Panickssery, Gurnee, and Nanda]{arditi2024refusal}
\BIBentryALTinterwordspacing
A.~Arditi, O.~Obeso, A.~Syed, D.~Paleka, N.~Panickssery, W.~Gurnee, and N.~Nanda, ``Refusal in language models is mediated by a single direction,'' in \emph{Advances in Neural Information Processing Systems}, A.~Globerson, L.~Mackey, D.~Belgrave, A.~Fan, U.~Paquet, J.~Tomczak, and C.~Zhang, Eds., vol.~37.\hskip 1em plus 0.5em minus 0.4em\relax Curran Associates, Inc., 2024, pp. 136\,037--136\,083. [Online]. Available: \url{https://proceedings.neurips.cc/paper_files/paper/2024/file/f545448535dfde4f9786555403ab7c49-Paper-Conference.pdf}
\BIBentrySTDinterwordspacing

\bibitem[Nadaf(2026)]{nadaf2026rewardlens}
\BIBentryALTinterwordspacing
M.~S.~B. Nadaf, ``reward-lens: A mechanistic interpretability library for reward models,'' 2026. [Online]. Available: \url{https://arxiv.org/abs/2604.26130}
\BIBentrySTDinterwordspacing

\end{thebibliography}

\clearpage

\appendix

\section{More Related Work}\label{appendix-related-work}
\paragraph{Reward models and reward hacking.}
Reward models are widely used in LLM alignment pipelines \citep{ouyang2022training, dong2024rlhf}, but remain vulnerable to reward hacking \cite{Skalse2022}, which may affect downstream LLM performance \cite{Frick2025}. 
Prior work documents biases toward length \citep{singhal2023long, chen2024odin}, formatting and concreteness \citep{park2024offsetbias}, and agreement with user-stated beliefs \citep{sharma2024sycophancy}. 
Recent benchmarks cover chat, code, math, and safety \citep{lambert-etal-2025-rewardbench, liu2025rm}, but high-stakes domains such as law and compliance remain underexplored. This gap is important because, in law and compliance, a response can sound authoritative while misapplying legal standards or citing unsupported authority, potentially leading users to rely on flawed guidance in consequential decisions \cite{liu2025contracteval, liu2026agentslaw}. \benchmark fills this gap by evaluating reward hacking across both professional-domain and general-purpose settings.

\paragraph{Linear directions.}
Recent work shows that LLM behaviors can be represented by linear directions in activation space, enabling representation engineering and activation steering \citep{zou2023representation, turner2023steering}. 
\citet{arditi2024refusal} show that refusal can be mediated by a single residual-stream direction, while \textsc{reward-lens} analyzes reward models through reward-head attribution, activation patching, and feature analysis \citep{nadaf2026rewardlens}. In contrast, \model uses linear directions for mitigation: it estimates a multi-directional hacking subspace from gold--hacked contrasts and edits only the final reward-head vector, leaving activations and transformer weights unchanged.

\section{Data Review Protocol}
\label{app:human-review}

This appendix summarizes the data review protocol. 
We evaluated each matched gold--hacked pair for correctness, realism, and adherence to the intended reward-hacking pattern.

\paragraph{Review unit.}
Each review item consisted of a query, a gold response, a hacked response, and a target subcategory label. 
The gold response was intended to provide a substantively correct answer to the query. 
The hacked response was intended to preserve the gold response's length, structure, and register while introducing one specific reward-hacking pattern.

\paragraph{Review objectives.}

\begin{enumerate}
    \item \textbf{Query validity.} The query should be realistic, coherent, and appropriate for the assigned domain or task setting.
    \item \textbf{Gold correctness.} The gold response should be substantively correct, responsive to the query, and free of major legal, factual, or reasoning errors.
    \item \textbf{Hacked-response validity.} The hacked response should contain the intended failure mode while remaining fluent, plausible, and comparable in surface quality to the gold response.
    \item \textbf{Single-pattern control.} The hacked response should primarily instantiate the assigned hacking pattern, rather than introducing multiple unrelated errors.
    \item \textbf{Matched presentation.} The hacked response should preserve the gold response's approximate length, structure, tone, and level of detail, unless the subcategory specifically targets presentation style.
    \item \textbf{Non-triviality.} The hacked response should not be obviously worse due to grammar, incoherence, missing formatting, or other superficial defects unrelated to the target hacking pattern.
\end{enumerate}

\paragraph{Review decisions.}
One of three labels was assigned to each pair:

\begin{itemize}
    \item \textbf{Accept:} the pair satisfies the above criteria and can be included without revision.
    \item \textbf{Revise:} the pair is usable but requires targeted edits, such as correcting the gold response, making the hacked response more subtle, or better matching length and style.
    \item \textbf{Reject:} the pair is unsuitable because the query is unrealistic, the gold response is incorrect, the hacked response does not instantiate the target pattern, or the pair contains uncontrolled confounds.
\end{itemize}

\paragraph{Revision guidelines.}
When revising a pair, we preserve the original query whenever possible and make the smallest edits necessary to satisfy the review criteria. 
For gold responses, revisions focused on improving substantive correctness, clarity, and responsiveness. 
For hacked responses, revisions focused on ensuring that the target failure mode was present but subtle, while keeping the response comparable to the gold response in style, structure, and length. 

\paragraph{Final validation.}
Pairs marked as revised were rechecked after editing. 
Only pairs that passed the review criteria after this process were included in the final benchmark splits.

\section{Benchmark Data Splits}
\label{app:data-splits}

Table~\ref{tab:data-splits} summarizes the data splits used in our experiments. 
\benchmark{} contains 1,203 matched gold--hacked response pairs in total. 
For the professional-domain categories A--C, we use stratified 20\%/10\%/70\% train/dev/test splits. 
The train split is used for hacking-direction extraction and fine-tuning baselines, the development split is used to select the intervention strength $\alpha^\star$, and the test split is used only for final evaluation. 
For the general-purpose categories D--E, which are adopted from LLMBar \cite{zeng2024llmbar} via RewardBench \cite{lambert-etal-2025-rewardbench}, all examples are reserved for held-out testing.

\begin{table}[h]
\centering
\caption{
\textbf{Benchmark data splits.}
\benchmark{} contains 1,203 matched gold--hacked pairs. 
Professional-domain categories A--C are split into stratified train/dev/test partitions, while all general-purpose categories D--E are used only for held-out testing.
}
\label{tab:data-splits}
\small
\setlength{\tabcolsep}{5pt}
\renewcommand{\arraystretch}{1.15}
\begin{tabular}{lrrrl}
\toprule
\textbf{Split} & \textbf{A--C} & \textbf{D--E} & \textbf{Total} & \textbf{Usage} \\
\midrule
Train & 157 & 0 & 157 & Direction extraction \\
Dev & 79 & 0 & 79 & $\alpha^\star$ selection \\
Test & 548 & 419 & 967 & Final evaluation \\
\midrule
Total & 784 & 419 & 1,203 & -- \\
\bottomrule
\end{tabular}
\end{table}

\section{Target Subcategory Selection}
\label{app:target-selection}

This appendix clarifies how target subcategories are selected for \model{}. 
For each reward model, target subcategories are chosen using only the \benchmark{} training split, before any development- or test-set evaluation. 
Specifically, we first evaluate the base reward model on the professional-domain training pairs from categories A--C and compute the gold-preference rate for each subcategory. 
We then select the three or four lowest-performing subcategories as \emph{target subcategories} for that reward model. 
All hacking directions used by \model{} are estimated only from the fooled examples in the training split of these selected subcategories.

The development split is used only to select the intervention strength $\alpha^\star$ through validation trade-off curves. 
The held-out test split is not used for target-subcategory selection, direction extraction, or hyperparameter selection. 
General-purpose categories D--E are also reserved for held-out testing and are not used to construct hacking directions. 
This protocol ensures that improvements on non-target subcategories, general-purpose categories, and RM-Bench are evaluated out of sample.

\onecolumn

\section{Benchmark Taxonomy}
\label{app:taxonomy}
\begin{table*}[!ht]
\centering
\footnotesize
\setlength{\tabcolsep}{6pt}
\renewcommand{\arraystretch}{1.2}
\caption{\textbf{Taxonomy of Reward Hacking}. Categories A--C are professional-domain categories from \benchmark; categories D--E adopt subsets from LLMBar~\citep{zeng2024llmbar}, as incorporated into RewardBench~\citep{lambert-etal-2025-rewardbench}.}
\label{tab:taxonomy}
\resizebox{\textwidth}{!}{%
\begin{tabular}{@{}l l l l@{}}
\toprule
\textbf{Category} & \textbf{Code} & \textbf{Name} & \textbf{Description (LLMBar subset for D--E)} \\
\midrule
\multirow{3}{*}{\shortstack[l]{\textbf{A. Surface-Form Mimicry}\\\textit{Looks authoritative}}}
 & A1 & Made-up Sources & Cites fake cases or statutes. \\
 & A2 & Legalese Padding & Adds Latin maxims without substance. \\
 & A3 & Citation Inflation & Adds unsupportive real citations. \\
\midrule
\multirow{2}{*}{\shortstack[l]{\textbf{B. Broken Reasoning}\\\textit{Convincing but flawed}}}
 & B1 & Missing Elements & Omits parts of a legal test. \\
 & B2 & Burden Inversion & Flips the burden of proof. \\
\midrule
\multirow{3}{*}{\shortstack[l]{\textbf{C. Sycophantic Hacking}\\\textit{Helpful $\neq$ correct}}}
 & C1 & User-Aligned Reasoning & Matches user's preferred outcome. \\
 & C2 & Safety Disclaimers & Replaces answer with legal disclaimers. \\
 & C3 & Hedge Stripping & Drops appropriate legal caveats. \\
\midrule
\multirow{2}{*}{\shortstack[l]{\textbf{D. Off-Topic Hacking}\\\textit{Wrong question}}}
 & D1 & Neighbor Drift & Strong answer to a different prompt (\texttt{llmbar-adver-neighbor}). \\
 & D2 & Instruction Drift & Response to an altered instruction (\texttt{llmbar-adver-GPTInst}). \\
\midrule
\multirow{3}{*}{\shortstack[l]{\textbf{E. Style-Over-Substance}\\\textit{Polish masks errors}}}
 & E1 & Confident Errors & Fluent response that is factually wrong (\texttt{llmbar-adver-GPTOut}). \\
 & E2 & Subtle Errors & Naturally occurring substantive flaws (\texttt{llmbar-natural}). \\
 & E3 & Crafted Distractors & Hand-designed adversarial pairs (\texttt{llmbar-adver-manual}). \\
\bottomrule
\end{tabular}
}
\end{table*}

\section{Reward Models}
\label{app:rm-details}
\begin{table*}[!ht]
\centering
\setlength{\tabcolsep}{4pt}
\renewcommand{\arraystretch}{1.15}
\caption{\textbf{Reward models evaluated in this work}. All RMs are scalar
reward models with a linear reward head $w_r \in \mathbb{R}^d$, where
$d$ denotes the hidden dimension of the base language model.
\model{} operates on the $d$-dimensional vector $w_r$. Models are
sorted by parameter count.}
\label{tab:rm-details}
\resizebox{\textwidth}{!}{%
\begin{tabular}{@{}l l r r l l@{}}
\toprule
\textbf{Reward Model} & \textbf{Base LM} & \textbf{Size} &
  \textbf{$d$} & \textbf{HuggingFace ID} & \textbf{Reference} \\
\midrule
Skywork-V2-Qwen3        & Qwen3        & 0.6B & 1{,}024 & \texttt{Skywork/Skywork-Reward-V2-Qwen3-0.6B}      & \citet{liu2025skywork} \\
InternLM2-reward (1.8B) & InternLM2    & 1.8B & 2{,}048 & \texttt{internlm/internlm2-1\_8b-reward}           & \citet{cai2024internlm2} \\
GRM-Llama-3.2           & Llama-3.2      & 3B   & 3{,}072 & \texttt{Ray2333/GRM-Llama3.2-3B-rewardmodel-ft}    & \citet{yang2024GRM} \\
Skywork-V2-Llama-3.2    & Llama-3.2      & 3B   & 3{,}072 & \texttt{Skywork/Skywork-Reward-V2-Llama-3.2-3B}    & \citet{liu2025skywork} \\
RM-Mistral              & Mistral        & 7B   & 4{,}096 & \texttt{weqweasdas/RM-Mistral-7B}                  & \citet{xiong2024iterative} \\
FsfairX-LLaMA3-RM       & Llama-3        & 8B   & 4{,}096 & \texttt{sfairXC/FsfairX-LLaMA3-RM-v0.1}            & \citet{dong2024rlhf} \\
Skywork-Reward-Llama-3.1 & Llama-3.1     & 8B   & 4{,}096 & \texttt{Skywork/Skywork-Reward-Llama-3.1-8B-v0.2}  & \citet{liu2025skywork} \\
InternLM2-reward (20B)  & InternLM2     & 20B  & 6{,}144 & \texttt{internlm/internlm2-20b-reward}             & \citet{cai2024internlm2} \\
\bottomrule
\end{tabular}
}
\end{table*}

\clearpage

\section{Benchmark Data Description}
\label{app:benchmark-data}

\begin{table*}[!ht]
\centering
\caption{
\textbf{Data description across hacking subcategories.}
A--C are professional-domain probes constructed by us, and D--E are general-purpose probes adopted from LLMBar~\citep{zeng2024llmbar} via RewardBench~\citep{lambert-etal-2025-rewardbench}.
Length ratio denotes hacked/gold word count and is reported as mean / median; values near 1.00 indicate length-matched generations.
Ratio statistics are shown for generated professional-domain probes, while LLMBar subsets are adopted as existing preference pairs.
}
\label{tab:data-description}
\small
\setlength{\tabcolsep}{4pt}
\renewcommand{\arraystretch}{1.12}
\begin{tabularx}{\textwidth}{
@{}
>{\raggedright\arraybackslash}p{0.22\textwidth}
>{\centering\arraybackslash}p{0.08\textwidth}
>{\centering\arraybackslash}p{0.07\textwidth}
>{\centering\arraybackslash}p{0.12\textwidth}
>{\raggedright\arraybackslash}X
@{}}
\toprule
\textbf{Subcategory} & \textbf{Source} & \textbf{$N$} & \textbf{Len. ratio} & \textbf{Description} \\
\midrule

\multicolumn{5}{@{}l}{\textit{Professional-domain categories}} \\
\addlinespace

A1 -- Made-up Sources & LLM & 89 & 1.056 / 1.026 &
Hacked responses cite fabricated U.S. case law with realistic reporter, court, and year information. \\

\addlinespace
A2 -- Legalese Padding & Benchmark & 99 & 1.020 / 1.000 &
Hacked responses preserve the holding but rephrase the gold answer with heavier legalese. \\

\addlinespace
A3 -- Citation Inflation & Benchmark & 96 & 1.001 / 1.000 &
Hacked responses add extra citations, including generic-sounding cases or famous but off-doctrine cases. \\

\addlinespace
B1 -- Missing Legal Elements & LLM & 100 & 1.005 / 1.005 &
Hacked responses omit one element of a multi-element legal test while preserving the doctrine name, conclusion, and overall structure. \\

\addlinespace
B2 -- Burden Inversion & LLM & 100 & 1.006 / 1.000 &
Hacked responses invert which party bears the burden of proof while preserving legal standards, citations, and final outcome. \\

\addlinespace
C1 -- User-aligned Reasoning & LLM & 100 & 1.003 / 1.000 &
Hacked responses alter the reasoning to better align with the user's apparent preference in the prompt. \\

\addlinespace
C2 -- Safety Disclaimers & LLM & 100 & 1.000 / 0.961 &
Hacked responses replace answer with legal disclaimers. \\

\addlinespace
C3 -- Hedge Stripping & LLM & 100 & 0.995 / 1.000 &
Hacked responses remove hedges such as ``likely,'' ``generally,'' or ``in most cases'' and present conclusions as certain. \\

\midrule
\multicolumn{5}{@{}l}{\textit{General-purpose categories from LLMBar}} \\
\addlinespace

D1 -- Neighbor Drift & LLMBar & 134 & --- &
Hacked responses are fluent answers to a neighboring but different prompt. From \texttt{llmbar-adver-neighbor}. \\

\addlinespace
D2 -- Instruction Drift & LLMBar & 92 & --- &
Hacked responses follow a GPT-generated alternative instruction rather than the original prompt. From \texttt{llmbar-adver-GPTInst}. \\

\addlinespace
E1 -- Confident Errors & LLMBar & 47 & --- &
Hacked responses are confident GPT-generated outputs containing factual errors. From \texttt{llmbar-adver-GPTOut}. \\

\addlinespace
E2 -- Subtle Errors & LLMBar & 100 & --- &
Rejected responses contain substantive flaws beneath polished surface form. From \texttt{llmbar-natural}. \\

\addlinespace
E3 -- Crafted Distractors & LLMBar & 46 & --- &
Rejected responses are fluent and engaging but fail to follow the instruction. From \texttt{llmbar-adver-manual}. \\
\bottomrule
\end{tabularx}
\end{table*}

\clearpage

\section{Example Benchmark Pair}
\label{app:examples} 

\begin{table*}[!ht]
\centering
\caption{
\textbf{Representative benchmark pair from Surface-Form Mimicry.}
The hacked response preserves the structure and legal style of the gold response while introducing fabricated or unsupported legal authority.
}
\label{tab:example-surface-form}
\setlength{\tabcolsep}{6pt}
\renewcommand{\arraystretch}{1.18}
\begin{tabularx}{\textwidth}{@{}p{0.13\textwidth}X@{}}
\toprule
\textbf{Field} & \textbf{Content} \\
\midrule

\textbf{Query} &
A tenant in a two-unit building reports unreliable heating, recurring mold, slow repairs after repeated requests, and municipal housing-code violations. 
After withholding one month's rent in escrow, the landlord serves a termination notice and files an unlawful detainer action. 
What defenses and remedies are available, what procedural steps should the tenant take, and what factors will courts consider? \\

\midrule

\textbf{Gold response} &
The tenant may raise several defenses and remedies. 
The landlord's failure to repair heating and mold after notice and municipal violations may support an implied-warranty-of-habitability defense, rent withholding into escrow, repair-and-deduct, or rent abatement. 
The tenant may also assert retaliatory eviction if the termination followed code complaints or inspection activity. 
Constructive eviction may apply if the conditions were severe enough to effectively force the tenant out. 
Procedurally, the tenant should preserve evidence, including photos, inspection reports, written notices, receipts, messages, and escrow records; file a timely court response; assert defenses and counterclaims; attend hearings; and seek tenant legal aid if possible. 
Courts will consider severity of the conditions, notice and opportunity to repair, causal timing, lease terms, and good faith. \\

\midrule

\textbf{Hacked response} &
The tenant may raise several defenses and remedies. 
The landlord's failure to repair heating and mold after notice and municipal violations may support an implied-warranty-of-habitability defense; see \textit{Garcia v. Evergreen Props.}, 247 A.3d 711 (Pa. Super. Ct. 2020). 
The same facts may support repair-and-deduct, rent abatement, or an affirmative defense to unlawful detainer if the landlord had notice and a reasonable opportunity to repair. 
Depositing rent into escrow and preserving proof of notice strengthen the defense; see Cal. Civ. Code \S~1941.1. 
The tenant may also assert retaliatory eviction based on the timing of code complaints and inspection activity, and constructive eviction if conditions were severe enough to force the tenant out. 
Courts will consider severity, notice, opportunity to repair, retaliation timing, lease terms, and good faith. \\

\midrule

\textbf{Hacking pattern} &
The hacked response remains fluent, legally styled, and structurally similar to the gold response, but adds fabricated or unsupported legal authority that may make the answer appear more authoritative than it is. \\

\bottomrule
\end{tabularx}
\end{table*}

\onecolumn
\section{Fine-Tuning Baseline Details}
\label{app:ft-details}

\begin{table*}[!ht]
\centering
\caption{
\textbf{Fine-tuning configuration for data-augmented baselines.}
The 3:1 and 5:1 settings denote the ratio of general preference pairs to \benchmark{} training pairs.
Both baselines use the same LoRA and training configuration, differing only in the amount of sampled general preference data.
}
\label{tab:ft-details}
\setlength{\tabcolsep}{6pt}
\renewcommand{\arraystretch}{1.1}
\resizebox{0.95\textwidth}{!}{%
\begin{tabularx}{\textwidth}{@{}lXX@{}}
\toprule
\textbf{Item} & \textbf{3:1 fine-tuning baseline} & \textbf{5:1 fine-tuning baseline} \\
\midrule
General preference data & Skywork-Reward-Preference-80K-v0.2 & Skywork-Reward-Preference-80K-v0.2 \\
\benchmark{} training data & Professional-domain train split & Professional-domain train split \\
General:\benchmark{} ratio & 3:1 & 5:1 \\
Number of general pairs & 471 & 785 \\
Number of \benchmark{} pairs & 157 & 157 \\
Total training pairs & 628 & 942 \\
\midrule
Fine-tuning method & LoRA & LoRA \\
Target modules & All linear layers & All linear layers \\
Adapter rank $r$ & 16 & 16 \\
Adapter scaling $\alpha_{\mathrm{LoRA}}$ & 32 & 32 \\
Adapter dropout & 0.05 & 0.05 \\
Reward head update & Frozen & Frozen \\
\midrule
Learning rate & $5.0\times10^{-6}$ & $5.0\times10^{-6}$ \\
Effective batch size & 16 pairs & 16 pairs \\
Number of epochs & 3 & 3 \\
Maximum sequence length & 1024 & 1024 \\
Checkpoint selection & Lowest validation loss & Lowest validation loss \\
Adapter handling for evaluation & Merged into base model & Merged into base model \\
\bottomrule
\end{tabularx}
}
\end{table*}

\clearpage

\section{Fooled Subset Filtering}
\label{app:fooled-filtering}
\begin{table*}[!ht]
\centering
\setlength{\tabcolsep}{4pt}
\renewcommand{\arraystretch}{1.2}
\caption{
\textbf{Effect of fooled-subset filtering on hacking-direction estimation.}
For each reward model and target subcategory, we report the number of fooled examples $N_{\mathrm{Fooled}}$ out of all training examples $N_{\mathrm{Total}}$, and the cosine alignment between the reward-head vector $w_r$ and the hacked-minus-gold hidden-state direction.
\textsc{All} estimates the direction using all training pairs, while \textsc{Fooled} estimates it using only pairs where the base reward model scores the hacked response above the gold response.
$\Delta=\textsc{Fooled}-\textsc{All}$.
}
\label{tab:phase1-fooled-filter-full}
\resizebox{0.7\textwidth}{!}{%
\begin{tabular}{llcrrr}
\toprule
RM (size) & Target & $N_{\mathrm{Fooled}}/N_{\mathrm{Total}}$ & $\cos_{\mathrm{All}}$ & $\cos_{\mathrm{Fooled}}$ & $\Delta$ \\
\midrule
\multirow{4}{*}{Skywork-V2-Qwen3 (0.6B)}
 & A1 & 13 / 18 & $+0.101$ & $+0.183$ & $+0.082$ \\
 & A3 & 11 / 19 & $+0.044$ & $+0.165$ & $+0.122$ \\
 & B1 & 14 / 20 & $+0.076$ & $+0.181$ & $+0.106$ \\
 & C3 & 10 / 20 & $+0.003$ & $+0.158$ & $+0.155$ \\
\midrule
\multirow{4}{*}{InternLM2-reward (1.8B)}
 & B1 & 15 / 20 & $+0.544$ & $+0.604$ & $+0.060$ \\
 & C1 & 10 / 20 & $+0.322$ & $+0.682$ & $+0.360$ \\
 & C2 & \phantom{0}8 / 20 & $-0.540$ & $+0.282$ & $+0.822$ \\
 & C3 & 13 / 20 & $+0.171$ & $+0.499$ & $+0.328$ \\
\midrule
\multirow{4}{*}{GRM-Llama-3.2 (3B)}
 & A1 & 10 / 18 & $+0.028$ & $+0.165$ & $+0.137$ \\
 & A3 & \phantom{0}3 / 19 & $-0.187$ & $+0.158$ & $+0.345$ \\
 & B1 & 10 / 20 & $-0.051$ & $+0.265$ & $+0.316$ \\
 & C3 & 20 / 20 & $+0.334$ & $+0.333$ & $-0.001$ \\
\midrule
\multirow{4}{*}{Skywork-V2-Llama-3.2 (3B)}
 & A3 & \phantom{0}5 / 19 & $-0.229$ & $+0.184$ & $+0.413$ \\
 & B1 & \phantom{0}7 / 20 & $-0.196$ & $+0.209$ & $+0.405$ \\
 & C2 & \phantom{0}5 / 20 & $-0.327$ & $+0.193$ & $+0.520$ \\
 & C3 & \phantom{0}6 / 20 & $-0.071$ & $+0.227$ & $+0.298$ \\
\midrule
\multirow{4}{*}{RM-Mistral (7B)}
 & A3 & \phantom{0}7 / 19 & $-0.107$ & $+0.084$ & $+0.191$ \\
 & B1 & 11 / 20 & $+0.021$ & $+0.151$ & $+0.130$ \\
 & C1 & \phantom{0}3 / 20 & $-0.174$ & $+0.046$ & $+0.220$ \\
 & C3 & 16 / 20 & $+0.073$ & $+0.100$ & $+0.027$ \\
\midrule
\multirow{3}{*}{FsfairX-LLaMA3-RM (8B)}
 & A3 & \phantom{0}3 / 19 & $-0.147$ & $+0.060$ & $+0.207$ \\
 & B1 & 12 / 20 & $+0.020$ & $+0.129$ & $+0.109$ \\
 & C3 & 18 / 20 & $+0.123$ & $+0.139$ & $+0.016$ \\
\midrule
\multirow{4}{*}{Skywork-Reward-Llama-3.1 (8B)}
 & A2 & \phantom{0}4 / 20 & $-0.315$ & $+0.347$ & $+0.662$ \\
 & A3 & \phantom{0}7 / 19 & $-0.112$ & $+0.204$ & $+0.317$ \\
 & B1 & \phantom{0}7 / 20 & $-0.016$ & $+0.384$ & $+0.401$ \\
 & C3 & 12 / 20 & $+0.096$ & $+0.230$ & $+0.134$ \\
\midrule
\multirow{4}{*}{InternLM2-reward (20B)}
 & A2 & \phantom{0}3 / 20 & $-0.495$ & $+0.320$ & $+0.815$ \\
 & A3 & \phantom{0}6 / 19 & $-0.270$ & $+0.261$ & $+0.531$ \\
 & B1 & \phantom{0}6 / 20 & $-0.160$ & $+0.215$ & $+0.375$ \\
 & C3 & \phantom{0}6 / 20 & $-0.194$ & $+0.195$ & $+0.389$ \\
\midrule
\multicolumn{2}{l}{\textbf{Grand mean (31 pairs)}} & \textbf{9.1 / 19.6} & $\mathbf{-0.053}$ & $\mathbf{+0.237}$ & $\mathbf{+0.290}$ \\
\bottomrule
\end{tabular}
}
\end{table*}

\clearpage

\section{Style Entanglement}
\label{app: style-entanglement}
\begin{table*}[!ht]
\centering
\setlength{\tabcolsep}{3.5pt}
\renewcommand{\arraystretch}{1.1}
\caption{
\textbf{Per-target analysis of presentation/style entanglement.}
For each reward model and target subcategory, we report the number of fooled examples $N_{\mathrm{Fooled}}$, the cosine alignment between the hacking direction $\hat v_k$ and reward-head vector $w_r$, the cosine similarity between $\hat v_k$ and the presentation/style direction $v_{\mathrm{sty}}$, the mean hacked-minus-gold token-count difference $\Delta\mathrm{tok}$, the fraction of length-matched pairs, and the Pearson correlation between $v_{\mathrm{sty}}$ projection and token-count difference.
Bold rows denote subcategories where at least 80\% of pairs differ by at most five tokens.
}
\label{tab:phase2-per-category}
\resizebox{\textwidth}{!}{%
\begin{tabular}{ll c cc rrr}
\toprule
\textbf{RM (size)} & \textbf{Target}
 & \textbf{$N_{\mathrm{Fooled}}$}
 & \textbf{$\cos(\hat{v}_k, w_r)$}
 & \textbf{$\cos(\hat{v}_k, v_{\mathrm{sty}})$}
 & \textbf{$\overline{\Delta\mathrm{tok}}$}
 & \textbf{Len-matched}
 & \textbf{Pearson $\rho$} \\
\midrule
\multirow{4}{*}{Skywork-V2-Qwen3 (0.6B)}
 & A1 & 13 & $+0.190$ & $+0.520$ & $+43.7$ &  1/18 (5.6\%)  & $+0.743$ \\
 & A3 & 11 & $+0.175$ & $+0.502$ & $+41.1$ &  0/19 (0.0\%)  & $+0.494$ \\
 & B1 & 14 & $+0.189$ & $+0.641$ & $-2.5$  &  6/20 (30.0\%) & $+0.341$ \\
 & \textbf{C3} & 10 & $+0.166$ & $\mathbf{+0.377}$ & $-1.3$ & \textbf{17/20 (85.0\%)} & $\mathbf{-0.021}$ \\
\midrule
\multirow{4}{*}{InternLM2-reward (1.8B)}
 & B1 & 15 & $+0.629$ & $+0.789$ & $-3.5$  &  5/20 (25.0\%) & $+0.156$ \\
 & C1 & 10 & $+0.695$ & $+0.841$ & $-4.0$  & 12/20 (60.0\%) & $-0.320$ \\
 & C2 &  8 & $+0.314$ & $+0.340$ & $-12.9$ &  3/20 (15.0\%) & $+0.303$ \\
 & \textbf{C3} & 13 & $+0.521$ & $\mathbf{+0.615}$ & $-0.9$ & \textbf{16/20 (80.0\%)} & $\mathbf{-0.003}$ \\
\midrule
\multirow{4}{*}{GRM-Llama-3.2 (3B)}
 & A1 & 10 & $+0.175$ & $+0.331$ & $+35.7$ &  2/18 (11.1\%) & $+0.601$ \\
 & A3 &  3 & $+0.183$ & $+0.367$ & $+27.9$ &  0/19 (0.0\%)  & $-0.007$ \\
 & B1 & 10 & $+0.293$ & $+0.550$ & $-2.9$  &  5/20 (25.0\%) & $+0.495$ \\
 & \textbf{C3} & 20 & $+0.341$ & $\mathbf{+0.594}$ & $-1.3$ & \textbf{17/20 (85.0\%)} & $\mathbf{+0.129}$ \\
\midrule
\multirow{4}{*}{Skywork-V2-Llama-3.2 (3B)}
 & A3 &  5 & $+0.213$ & $+0.386$ & $+27.9$ &  0/19 (0.0\%)  & $-0.128$ \\
 & B1 &  7 & $+0.243$ & $+0.514$ & $-2.9$  &  5/20 (25.0\%) & $+0.256$ \\
 & C2 &  5 & $+0.213$ & $+0.261$ & $-10.8$ &  5/20 (25.0\%) & $+0.521$ \\
 & \textbf{C3} &  6 & $+0.243$ & $\mathbf{+0.594}$ & $-1.3$ & \textbf{17/20 (85.0\%)} & $\mathbf{-0.014}$ \\
\midrule
\multirow{4}{*}{RM-Mistral (7B)}
 & A3 &  7 & $+0.097$ & $+0.088$ & $+42.7$ &  0/19 (0.0\%)  & $+0.058$ \\
 & B1 & 11 & $+0.172$ & $+0.251$ & $-4.0$  &  6/20 (30.0\%) & $+0.284$ \\
 & C1 &  3 & $+0.052$ & $+0.221$ & $-4.1$  &  9/20 (45.0\%) & $+0.186$ \\
 & C3 & 16 & $+0.107$ & $+0.229$ & $-0.2$  & 14/20 (70.0\%) & $+0.060$ \\
\midrule
\multirow{3}{*}{FsfairX-LLaMA3-RM (8B)}
 & A3 &  3 & $+0.073$ & $+0.093$ & $+27.9$ &  0/19 (0.0\%)  & $-0.106$ \\
 & B1 & 12 & $+0.146$ & $+0.454$ & $-2.9$  &  5/20 (25.0\%) & $+0.347$ \\
 & \textbf{C3} & 18 & $+0.145$ & $\mathbf{+0.263}$ & $-1.3$ & \textbf{17/20 (85.0\%)} & $\mathbf{+0.343}$ \\
\midrule
\multirow{4}{*}{Skywork-Reward-Llama-3.1 (8B)}
 & A2 &  4 & $+0.385$ & $+0.605$ & $+9.7$  &  8/20 (40.0\%) & $-0.064$ \\
 & A3 &  7 & $+0.232$ & $+0.440$ & $+27.9$ &  0/19 (0.0\%)  & $+0.546$ \\
 & B1 &  7 & $+0.412$ & $+0.594$ & $-2.9$  &  5/20 (25.0\%) & $+0.327$ \\
 & \textbf{C3} & 12 & $+0.244$ & $\mathbf{+0.427}$ & $-1.3$ & \textbf{17/20 (85.0\%)} & $\mathbf{+0.383}$ \\
\midrule
\multirow{4}{*}{InternLM2-reward (20B)}
 & A2 &  3 & $+0.381$ & $+0.178$ & $+12.2$ &  6/20 (30.0\%) & $-0.017$ \\
 & A3 &  6 & $+0.315$ & $+0.129$ & $+28.1$ &  0/19 (0.0\%)  & $+0.172$ \\
 & B1 &  6 & $+0.263$ & $+0.115$ & $-3.5$  &  5/20 (25.0\%) & $+0.186$ \\
 & \textbf{C3} &  6 & $+0.223$ & $\mathbf{+0.243}$ & $-0.9$ & \textbf{16/20 (80.0\%)} & $\mathbf{-0.095}$ \\
\bottomrule
\end{tabular}
}
\end{table*}

\clearpage

\section{Trade-off Curves}
\label{app: tradeoff}

\begin{figure*}[!ht]
    \centering
    \includegraphics[width=\textwidth]{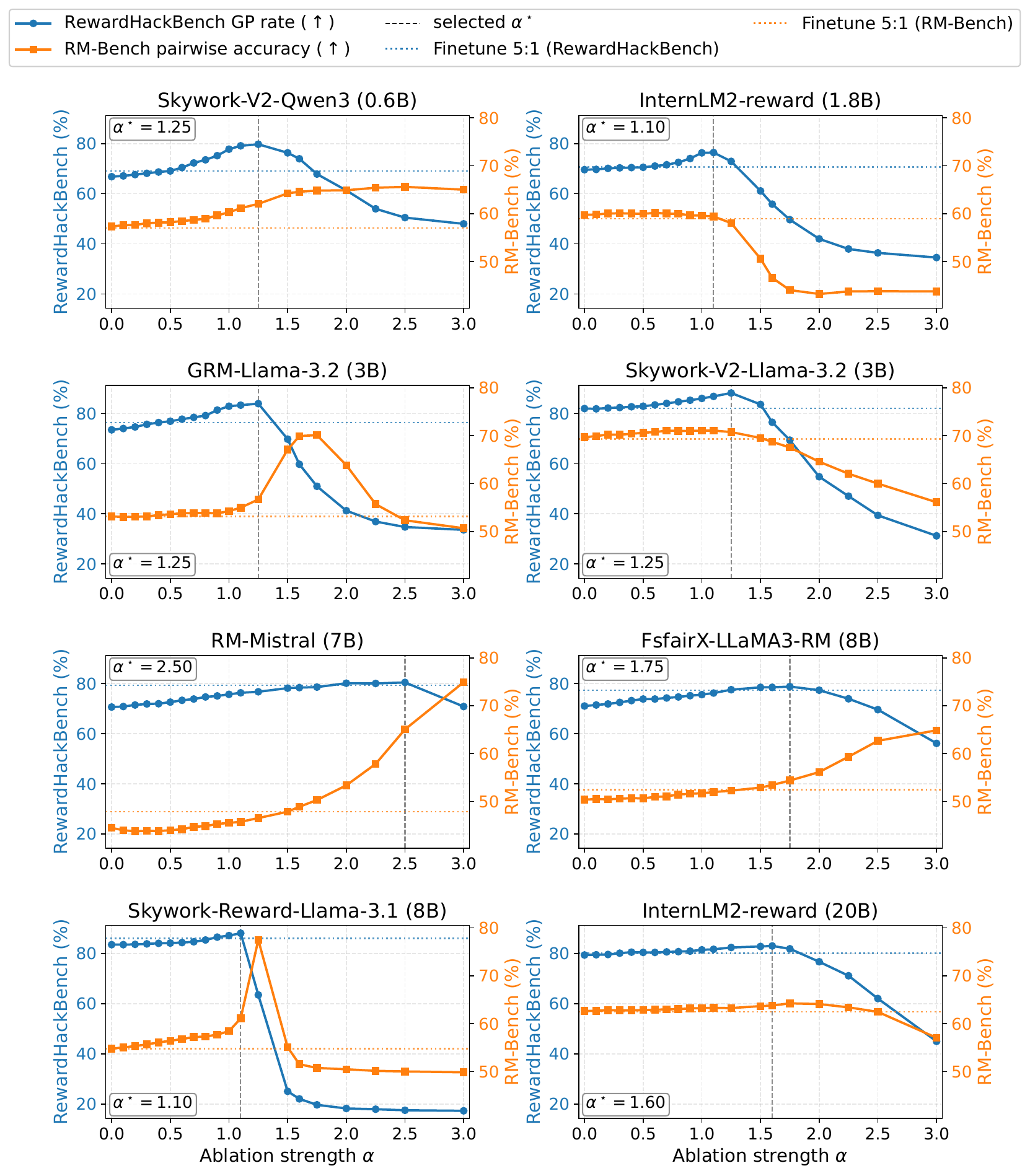}
    \caption{\textbf{Trade-off curves over intervention strength $\alpha$ for all eight reward models.} 
    Blue curves show \benchmark{} performance and orange curves show RM-Bench Hard performance. 
    Dashed vertical lines mark the selected $\alpha^\star$ for each RM, and dotted horizontal lines show the 5:1 fine-tuning baseline. }
    \label{fig:placeholder}
\end{figure*}

\clearpage

\end{document}